\documentclass[a4paper,12pt,oneside]{book}
\usepackage[utf8]{inputenc}
\frontmatter
\usepackage{graphicx}
\usepackage[]{footmisc}
\usepackage{listing}
\usepackage{makecell}
\graphicspath{ {images/} }
\usepackage{float}
\usepackage{hyperref}
\usepackage{geometry}
\usepackage{datetime}
\usepackage{lscape}
\usepackage{amsmath}
\usepackage{amssymb}
\usepackage{threeparttable}
\usepackage{booktabs}
\usepackage{threeparttable}
\usepackage{multirow}
\usepackage{algorithm,algpseudocode}
\usepackage{tikz}

\newdateformat{monthdayyeardate}{\monthname[\THEMONTH]~\THEDAY, \THEYEAR}
\usepackage{setspace}
\usepackage{blindtext}
\hypersetup{
    colorlinks=true,
    linkcolor=black,
    filecolor=magenta,      
    urlcolor=blue,
    citecolor=blue,
    pdftitle={MTech-Thesis-21MCMT02}
}
\title{MTech-Thesis}
\author{Arkaprabha Basu}
\date{June 2023}

\begin{document}
\begin{titlepage}
\begin{center}
    
    \Huge
    \textbf{Revealing the Ancient Beauty: Digital Reconstruction of Temple Tiles using Computer Vision}
    
    \vfill
    
    \small
    A Dissertation Submitted to the University of Hyderabad in Partial Fulfillment of the Degree of
        
    \vfill
    
    \Large
    Master of Technology\\
    \small
    in\\
    \large
    Computer Science
    
    \vspace{1cm}
    
    \small
    by\\
    \textbf{Arkaprabha Basu}\\
    21MCMT02
    
    \vfill
    
    \includegraphics[scale=1.7 ]{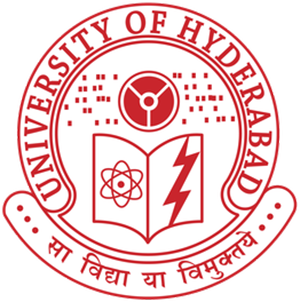}
    
    \vfill
    
    \large
    School of Computer and Information Science\\
    \textbf{University of Hyderabad}\\
    \small
    Gachibowli, Hyderabad - 500 046\\
    Telangana, India\\
    
    \vfill
    
    \textbf{\monthdayyeardate\today}

\end{center}
\end{titlepage} 

\vspace{-4cm}
\chapter*{\centering CERTIFICATE}
This is to certify that the dissertation titled, \textbf{“Revealing the Ancient Beauty: Digital Reconstruction of Temple Tiles using Computer Vision”} submitted by \textbf{Arkaprabha Basu}, Registration number: \textbf{21MCMT02}, in partial fulfillment of the requirements for the award of Master of Technology in Computer Science is a bonafide work carried out by him under my supervision and guidance.

I am absolutely certain that this dissertation is an exceptional example of intellectual innovation, representing a unique and original contribution that has never before been submitted, in whole or in part, to this esteemed institution or to any other accredited university or educational institution in pursuit of any degree or diploma.
\vfill
\vspace{3cm}







\begin{figure}[htbp]
  \centering
  \begin{minipage}[t]{0.3\linewidth}
    \centering
    \textbf{Prof. Chakravarthy Bhagvati}\\
    (Internal Supervisor)\\
    School of Computer and Information Sciences,\\
    University of Hyderabad
  \end{minipage}%
  \hfill
  \begin{minipage}[t]{0.3\linewidth}
    \centering
    \textbf{Prof. Swagatam Das}\\
    (External Supervisor)\\
    Electronics and Communication Sciences Unit, Indian Statistical Institute
  \end{minipage}%
  \hfill
  \begin{minipage}[t]{0.3\linewidth}
    \centering
    \textbf{Prof. Atul Negi}\\
    (Dean)\\
    School of Computer and Information Sciences,\\
    University of Hyderabad
  \end{minipage}%
\end{figure}
\vspace{1.5in} 
\begin{center}
\textbf{\huge Internship Certificate}
\end{center}
\noindent This is to certify that \textbf{Mr. Arkaprabha Basu} has successfully completed an internship project entitled \textbf{``Revealing the Ancient Beauty: Digital Reconstruction of Temple Tiles using Computer Vision"} at \textbf{Electronics and Communication Sciences Unit, Indian Statistical Institute, Kolkata} from June 2022.

During the internship period, Mr. Arkaprabha Basu exhibited exemplary dedication, commitment, and technical skills in carrying out the assigned tasks related to the project. His contributions were instrumental in the successful implementation of the project objectives.

I express my sincere gratitude for his valuable contributions to the success of the project and wish him the very best in his future academic and professional pursuits.
\vspace{3cm}
\begin{flushright}
\begin{minipage}[t]{0.3\linewidth}
    \centering
    \textbf{Prof. Swagatam Das}\\
    Electronics and Communication Sciences Unit,\\ Indian Statistical Institute, Kolkata
  \end{minipage}%
\end{flushright}

\chapter*{\centering DECLARATION}
I, Arkaprabha Basu, hereby declare that this dissertation titled, \textbf{“Revealing the Ancient Beauty: Digital Reconstruction of Temple Tiles using Computer Vision”}, submitted by \textbf{me} under the guidance and supervision of \textbf{Prof. Chakravarthy Bhagvati} and \textbf{Prof. Swagatam Das} is a bonafide work which is also free from plagiarism. I also declare that it has not been submitted previously in part or in full to this University or other University or Institution for the award of any degree or diploma. I hereby agree that my dissertation can be deposited in Shodhganga/INFLIBNET.

A report on plagiarism statistics from the University Librarian / iThenticate is enclosed.\\

\textbf{Date:} \hspace{6.15cm} \textbf{Name:} Arkaprabha Basu

\hspace{7.5cm} \textbf{Registration No.:} 21MCMT02
\vspace{1em}

\begin{center}
    //Countersigned//
\end{center}

Signature of the Supervisor(s):

\vspace{2.5cm}

(Prof. Chakravarthy Bhagvati)

\vspace{2.5cm}

(Prof. Swagatam Das) 
\chapter*{Acknowledgements}
I am filled with immense pleasure as I express my heartfelt gratitude to those individuals who have made the completion of this project possible. Foremost among them is \textbf{Prof. Swagatam Das}, whose invaluable guidance and provision of an exceptional research-friendly laboratory have been nothing short of extraordinary.  His mentorship and expertise have been pivotal in shaping the trajectory of this endeavor, and I am truly indebted to him. I would also like to extend my sincere appreciation to \textbf{Prof. Chakravarthy Bhagvati} for his unwavering support throughout this journey. His unwavering support and encouragement have been crucial in helping me overcome obstacles and reach significant goals.

\noindent My outstanding lab colleagues at the Indian Statistical Institute, Kolkata have provided me with priceless feedback, which has been a tremendous source of inspiration and personal development. My strategy has continuously improved as a result of their intelligent examination and perceptive perspective mapping, producing excellent results. They have truly been a blessing, and I am incredibly appreciative of what they have contributed.

\noindent In addition, I must express my sincere gratitude to the \textbf{Department of Science and Technology, India}, whose financial backing made it possible to cover the project's costs. 

\noindent Indeed, the successful completion of this endeavour is the result of the interaction of all of these factors, as well as the supportive atmosphere in which I have had the honour to work. I am incredibly appreciative of how these things came together to bring this project to a successful end.

\vspace{1cm}
\hspace{10cm} \textbf{Arkaprabha Basu}
\chapter*{Abstract}
Modern digitised approaches have dramatically changed the preservation and restoration of cultural treasures, integrating computer scientists into multidisciplinary projects with ease. This study scrutinises how contemporary digitised methods have revolutionised the preservation and restoration of cultural assets. Machine learning, deep learning, and computer vision techniques have revolutionised developing sectors like 3D reconstruction, picture inpainting, IoT-based methods, genetic algorithms, and image processing with the integration of computer scientists into multidisciplinary initiatives. We suggest three cutting-edge techniques in recognition of the special qualities of Indian monuments, which are famous for their architectural skill and aesthetic appeal. First is the Fractal Convolution methodology, a segmentation method based on image processing that successfully reveals subtle architectural patterns within these irreplaceable cultural buildings. The second is a revolutionary Self-Sensitive Tile Filling (SSTF) method created especially for West Bengal's mesmerising Bankura Terracotta Temples with a brand-new data augmentation method called MosaicSlice on the third. Furthermore, we delve deeper into the Super Resolution strategy to upscale the images without losing significant amount of quality. Our methods allow for the development of seamless region-filling and highly detailed tiles while maintaining authenticity using a novel data augmentation strategy within affordable costs introducing automation. By providing effective solutions that preserve the delicate balance between tradition and innovation, this study improves the subject and eventually ensures unrivalled efficiency and aesthetic excellence in cultural heritage protection. The suggested approaches advance the field into an era of unmatched efficiency and aesthetic quality while carefully upholding the delicate equilibrium between tradition and innovation.


\tableofcontents
\listoffigures
\listoftables
\chapter{List of acronyms}
\begin{center}
        \begin{table}[htb]
            \renewcommand{\arraystretch}{1.3}
            \centering
            \begin{tabular}{p{3cm} | p{12cm}}
            \hline
                SSTF & Self Sensitive Tile Filling \\
                YOLO & You Only Look Once \\
                SR & Super Resolution \\
                LR & Low Resolution \\
                HR & High Resolution \\
                VAE & Variational Autoencoder \\
                GAN & Generative Advarsarial Network \\
                DCGAN & Deep Convolutional Generative Adversarial Network \\
                CGAN & Conditional Generative Adversarial Network \\ 
                ProGAN & Progressive Growing Generative Adversarial Network \\
                SRGAN & Super Resolution Generative Adversarial Network \\
                ESRGAN & Enhanced Super Resolution Generative Adversarial Network \\
                BSRGAN & Blind Super Resolution Generative Adversarial Network \\
                ProTilesGAN & Progressive Tiles based Generative Adversarial Network \\
                Stage-Pro-SR & Progressive Stage Super Resolution \\
                WGAN & Wasserstein Generative Adversarial Network\\
                PH & Photogrammetry\\
                GB & GigaByte\\
                FD & Fractal Dimension\\
                FC & Fractal Convolution\\
                CNN & Convolutional Neural Network\\
                \hline
            \end{tabular}
            \caption{List of acronyms}
            \label{table:acronyms1}
        \end{table}
\end{center}
\chapter{List of acronyms}
\begin{center}    
            \begin{table}[htb]
            \renewcommand{\arraystretch}{1.3}
            \centering
            \begin{tabular}{p{3cm} | p{12cm}}
            \hline
                ML & Machine Learning\\
                DL & Deep Learning\\
                RGB & Red Green Blue\\
                ICP & Iterative Closest Point\\
                FSSR & Floating Scale Surface Reconstruction\\
                LST & Large Scale Texturing\\
                MVS & Multi View Stereo\\
                LIOP & Local Intensity Order Pattern\\
                UAS & Unmanned Aerial System\\
                AR & Augmented Reality\\
                GradCAM & Gradient Class Activation Map\\
                SSD & Single Shot Multibox Detector\\
                MAP & Mean Average Precision\\
                CSP & Cross Stage Partial\\
                VOC & Visual Object Class\\
                COCO & Common Object in Context\\
                NAFSSR & Nonlinear Activation Free Stereo Super Resolution\\
                HAT & Hybrid Attention Transformer\\
                JS & Jensen Shannon Divergence\\
                KL & Kullback Leibler Divergence\\
                FID & Frechet Inception Distance\\
                SSIM & Structural Similarity Index Metrics\\
            \hline
            \end{tabular}
            \caption{List of acronyms}
            \label{table:acronyms2}
        \end{table}
            \end{center} 
\newpage
\mainmatter

\chapter{Introduction}
\label{chapter:intro}
In recent years, the preservation of cultural heritages has become a compelling goal. Like a tapestry of tiles, these architectural marvels have elaborate shapes and patterns engraved into their façade. Notably, the magnificent mosques, madrasahs, tomb complexes, and mausoleums of Central Asian Timurid architecture serve as a spectacular example of this creativity. Timur, the founder of the Timurid empire, made a lasting impression on the territories he conquered, which included Iran, Iraq, Syria, the northern shores of the Black Sea, Anatolia, and the legendary Silk Road that linked China and Europe.

The Temple of Apollo at Syracuse, which is embellished with beautiful inscriptions that provide light on the beginnings of this ancient Greek city, is one outstanding example of architectural magnificence. Beyond buildings, artistic legacy includes respected monuments like the Holy Christ of Blood, a mystical and important historical figure in Spain. Another stunning illustration is the Neptune monument in Bologna, which was built in 1565 to oppose the strength of the newly enthroned Pope Pius IV. It is a lively representation of power and authority. The Baroque period's multicoloured wooden splendour known as the Holy Christ of Blood has significant archaeological and iconographic value.

\begin{figure}[!ht]
\label{extent}
	\centering
		\includegraphics[width=.7\linewidth]{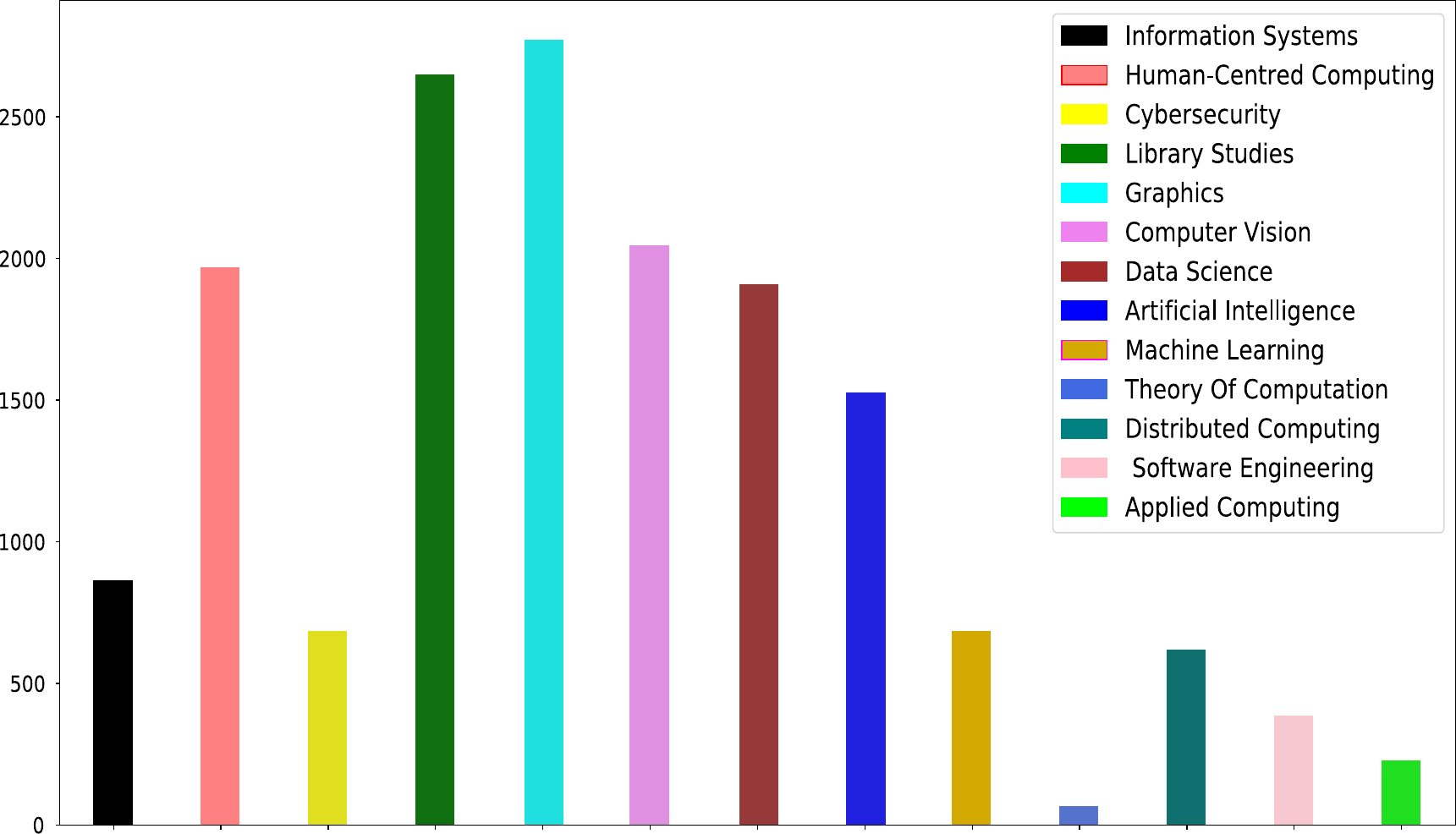}	
		\caption{Current Research Trend in Heritage Restoration and Preservation}
\end{figure}

However, as we go towards a digital future ruled by materialistic interests, these revered structures progressively disappear into obscurity, losing their cultural primacy over time. Fortunately, computer scientists are paying more and more attention to the field of heritage restoration and are utilising the power of computer science to revitalise these ancient wonders. Figure \ref{extent}, taken from the immense dimensions of web-based knowledge, gives an idea of the vast scope of this developing area and arouses the interest of computer scientists all around the world. Figure \ref{extent} taken from online source\footnote{\url{https://app.dimensions.ai/}} highlights various developing sites of computer science that are now attracting more computer scientists.
\section{Motivation}
\label{chapter:motivation}
Historic buildings and artefacts are priceless pieces of cultural heritage that must be maintained. Self-Sensitive Tile Filling (SSTF), an automated method, provides creative answers for accurate and efficient reconstruction by bridging the past and present while preserving historical accuracy and cultural appreciation.

\textbf{Preservation of Historical Artifacts:} Historic objects and buildings are priceless pieces of cultural heritage that must be safeguarded for future generations. However, erosion and deterioration brought on by the passage of time might result in the loss of important historical data. It is crucial to create methods that can successfully conserve and rebuild these artefacts.

\textbf{Limitations of Manual Reconstruction:} Traditional manual reconstruction techniques take a lot of time and are open to different interpretations. Manual reconstruction is difficult because of how detailed and complex cultural places and artefacts are, especially when working on large-scale projects. Automated methods are required in order to speed up the reconstruction process, guarantee accuracy, and save time and labour.

\textbf{Advancements in Automated Approaches:} To get around the drawbacks of human reconstruction, automated methods have arisen as creative solutions, such Self-Sensitive Tile Filling (SSTF). These methods automate the process of artefact reconstruction, producing more effective and precise results. They make use of computational algorithms, computer vision techniques, and machine learning.

\textbf{Self-Sensitive Tile Filling (SSTF):} SSTF focuses on intelligently analysing and matching tiles to patch up any gaps or damaged regions in an artefact. This method ensures a smooth integration into the overall reconstruction by accounting for the distinctive qualities and context of each tile. To find potential replacement tiles that preserve the artifact's historical accuracy and aesthetic coherence, SSTF uses self-sensitivity approaches.

\textbf{Enhancing Reconstruction Accuracy:} By increasing the precision of the reconstruction process, SSTF helps to preserve historical artefacts. SSTF makes sure that the filled portions perfectly integrate with the surrounding elements, maintaining the artifact's original aesthetics and finer nuances by utilising self-sensitivity and intelligent tile matching.

\textbf{Streamlining Reconstruction Workflow:} By requiring less manual labour, automated methods like SSTF simplify the reconstruction operation. SSTF automates the process by analysing the available tiles intelligently and choosing the best fit for each empty space, as opposed to picking and arranging each tile by hand. Time is saved, human error is reduced, and more accurate and efficient reconstructions are produced as a result.

\textbf{Historical Accuracy and Cultural Appreciation:} The use of automated methods like SSTF not only helps to preserve historical artefacts but also fosters a greater respect for and understanding of our cultural heritage. SSTF assists in revealing the original form and intricate details of the artefact by precisely rebuilding damaged or missing regions, giving us important new insights into the object's historical context and value.

\textbf{Bridging the Past and Present:} SSTF bridges the gap between the past and the present by combining automation and historical preservation. SSTF gives us the opportunity to experience and appreciate the cultural legacy that our predecessors left us by accurately and meticulously reproducing artefacts. The narrative and historical value of these artefacts are maintained through this procedure and handed on to subsequent generations.
\section{Contributions}    
\label{chapter:contributions}
Our suggested methodology offers a number of theoretical benefits and automation potential in the field of temple tile reconstruction by merging these cutting-edge computer vision techniques. First off, YOLOV8's automated detection of damaged tiles decreases the amount of manual work needed to determine the extent of the damage. Additionally, by combining ProTilesGAN and MosaicSlice, varied and contextually appropriate tile variations can be produced, boosting the depth and authenticity of the recreated tiles. Last but not least, StageWise Super Resolution is used to ensure that the replacement tiles maintain fine details and a high level of visual fidelity, resulting in a visually flawless integration with the undamaged tiles.
\section{Summary}
\label{chapter:summary}
We hope to streamline the repair of temple tiles using this all-encompassing technique, making it more effective, precise, and available. Our method preserves the exquisite beauty of temple building for future generations to appreciate and adore, which not only saves time and work but also safeguards the cultural legacy.
\section{Outline}
\label{chapter:outline}
\begin{figure*}[!htb]
\label{survayintro}
	\centering
		\includegraphics[width=\linewidth]{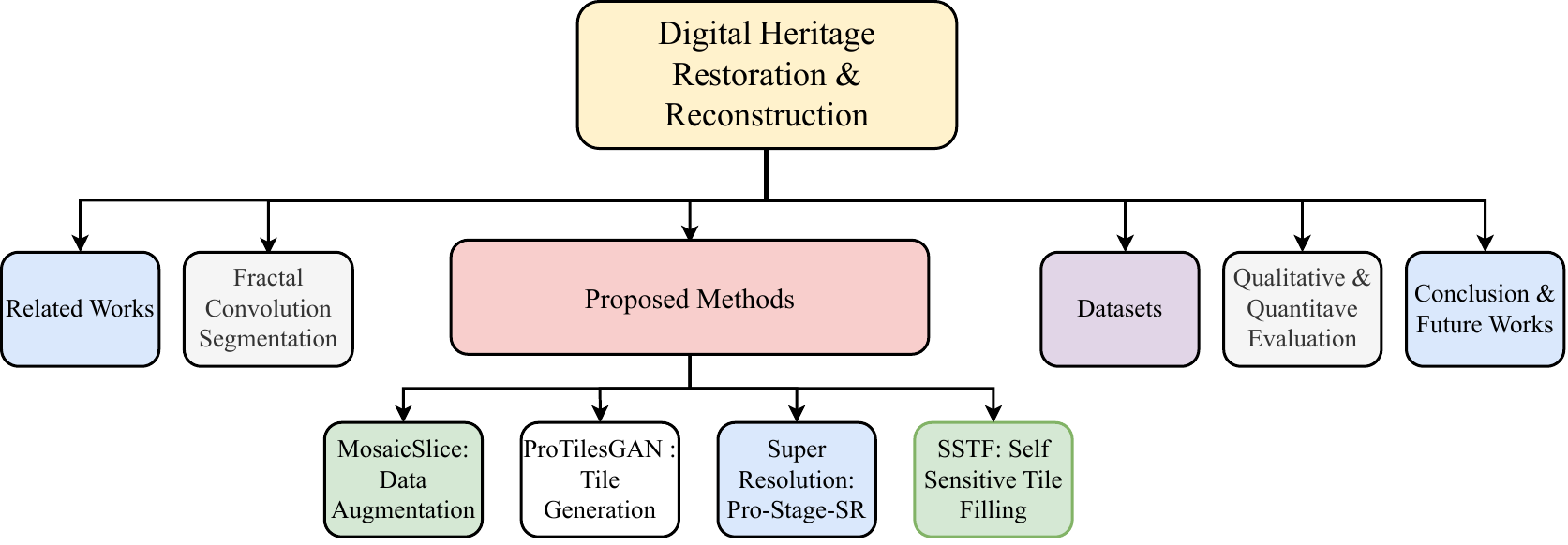}	
		\caption{Pictorial Recap of Subtopics}
\end{figure*}
\begin{itemize}
\item This fascinating chapter \ref{chapter:related_work} embarks on an illuminating journey into the complex world of top-notch research in this topic. We travel through history with a firm emphasis on specifics, investigating a wide range of study projects. Our journey focuses on cutting-edge automative ways that leverage the power of machine and deep learning, creative data collection methods, sophisticated 3D reconstruction techniques, and a variety of other ground-breaking technologies. These initiatives work together to push the limits of knowledge even further, ushering in a new era of invention and discovery.
\item In chapter \ref{ref:fc}, excluding lower regions, we introduce a novel fractal convolution method to segment rich and high details from heritage buildings. In order to compare whether this strategy truly outperforms the original data, we also test it using a conventional classifier network.
\item In chapter \ref{ref:mosiacslice}, we make our second contribution to this topic by presenting a cutting-edge data augmentation approach that will enable us to create deep learning networks with improved functionality.
\item We offer a novel GAN architecture inspired by the fundamental ProGAN in Chapter \ref{ref:protilesgan}, which combines new interpolation techniques in place of closest neighbour interpolation and an upgrade to the basic ProGAN's equalised learning rate.
\item Super resolution is put into force. In order to achieve higher performance, we stage out the capabilities of the standard SRGAN in our proposal for chapter \ref{ref:sr} to provide a novel super resolution GAN architecture for the purpose of Super Resolution.
\item In chapter \ref{ref:sstf}, we proudly present our final method, SSTF, as we come to the end of our study trip. In order to fix defective tiles inside temples, SSTF adopts a stage-wise approach, drawing inspiration from the wide tapestry of inventive worlds we have explored. We start a transformative project to give these sacred spaces new life, guided by the powerful capabilities of ProTilesGAN and Stage-SR. ProTilesGAN creativity and Stage-SR accuracy are skillfully combined by SSTF, creating a restorative symphony that restores the treasured structures' once-glowing brilliance.
\item Our research begins with three distinct datasets in the section \ref{ref:dataset}, which are then used to conduct experiments in the chapters \ref{ref:fc} \ref{ref:mosiacslice} \ref{ref:protilesgan}. In section \ref{ref:exp}, experiments were conducted using various qualitative and quantitative approaches.
\item In section \ref{ref:conclusion}, we draw a deep, profound conclusion to our research with the hope of preserving some astounding future achievements.
\end{itemize} 
\chapter{Related Work}
\label{chapter:related_work}
\section{Data Acquisition from Heritage Sites}
\subsection{Photogrammetry and 3D Scanning}
Photogrammetry (PH), also referred to as "Die Photometrographie," is a technique that makes it possible to gather accurate data about an object by measuring different aspects utilising electromagnetic radiant images, recording, and photo-taking \cite{basu_survay_digitalheritage2023}. It provides the theoretical underpinning for measuring from photos, especially when creating digital 3D models. The following steps are involved in photogrammetry-based data collection:
\begin{itemize}
    \item \textbf{Acquisition of images:} The core idea behind photogrammetry is that images serve as the main data source for creating 3D models. These photos should be taken using a calibrated camera and enough overlap to guarantee accurate reconstruction. According to \cite{basu_survay_digitalheritage2023}, successful coverage of the survey area and thorough picture collection depend on proper planning.
    \item \textbf{Preprocessing an image:} The obtained photos are preprocessed before analysis to remove any distortions that can jeopardise the correctness of the resulting 3D model.
    \item \textbf{Image Matching:} A crucial phase in photogrammetry is image matching, which involves finding traits that are shared by the pictures and creating correspondences between them. This procedure is essential for producing accurate and trustworthy 3D models.
    \item \textbf{Point Cloud Extraction:} The production of points After correspondences have been established, a point cloud is created using triangulation techniques (\cite{basu_survay_digitalheritage2023}). The 3D coordinates of the related features seen in the pictures are represented by this point cloud.
    \item \textbf{Surface Generation:} The surface generation is The point cloud is used as the starting point for creating a surface using interpolation, meshing, or contouring methods.
\end{itemize}
Through a series of clear stages, photogrammetry makes it easier to extract useful information from images, enabling the development of precise 3D models.

\subsection{Time Lapse imaging with rephotography}
A beautiful type of rephotography called time-lapse imaging reveals the temporal transformation of certain things. It plays a significant part in understanding the slow degradation of the stone curvatures embellishing architectural wonders and monuments within the context of digital heritage data collecting. In addition to acting as a catalyst for physical terrain restoration, its innovative use offers priceless insights into the underlying causes of the delicate flaking of particular places. Utilising the artistry of ultraviolet photography, X-ray radiography, and infrared photography to detect imperceptible damage eluding the human visual spectrum, this technique embraces various photographic modalities, skillfully capturing the subject at precise intervals. Interestingly, time-lapse imaging crosses disciplinary borders and is remarkably useful in the field of coal mining exploration, where it reveals the mysterious depths of old gallery mines. Furthermore, it inspires ground-breaking investigations based on careful segmentation in the field of bacterial cell track analysis \cite{basu_survay_digitalheritage2023}. By masterfully utilising this sublime imaging technique, fast-forward videos that vividly depict the episodic oscillations coursing through the visible object's essence and weave a timeless narrative are created.

\begin{figure}
    \centering
    \includegraphics[width=0.8\linewidth]{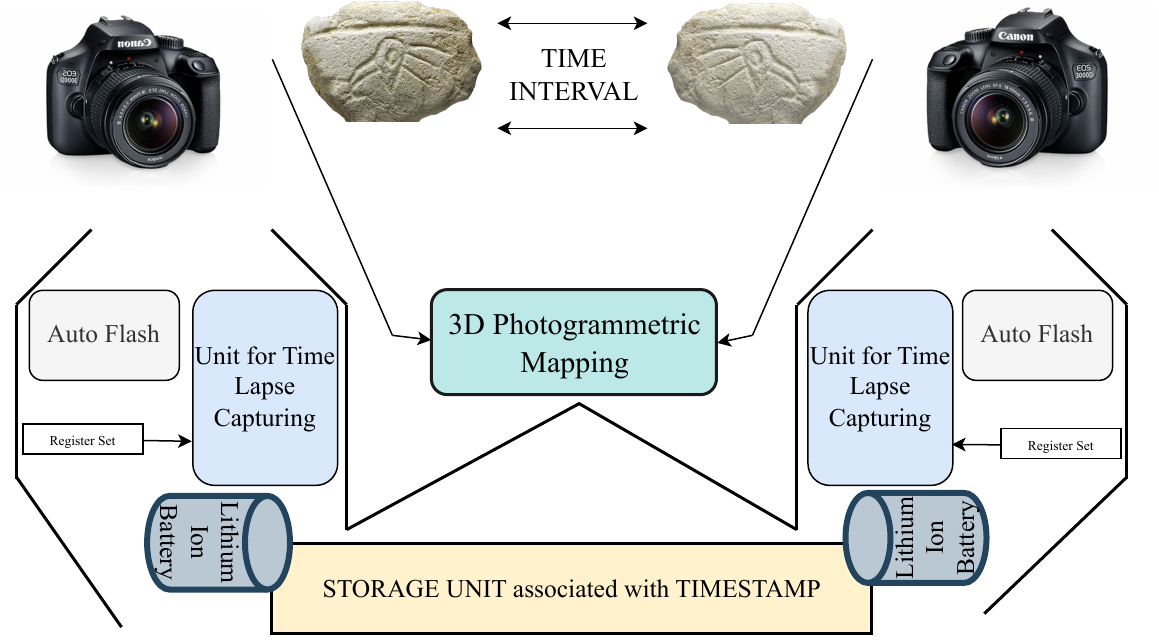}
    \caption{Time Lapse Imaging}
    \label{fig:time_lapse_imaging}
\end{figure}

\textbf{Analysis of Magnesium Limestone Flaking:}\\
Time-lapse imaging is used mentioned in \cite{basu_survay_digitalheritage2023} to look into the decomposition of magnesium limestones in historical places. According to the study, environmental conditions like strong winds or synchronised surface changes among nearby stone blocks are to blame for the episodic loss of stones. By adding terrain structure analysis, macro photography, and interval timers, the authors improve on conventional time-lapse photography. Every three months, a normal seven-pixel camera with 4GB of memory takes three hours of high-resolution photos. Adobe Photoshop is used for colour correction to maintain uniform lighting and exposure. For visual loss verification, the photos are then combined into an HD video using Apple Quicktime software. Finally, Photoshop is used to quantify error situations such surface alterations and edge variations. Figure \ref{fig:time_lapse_imaging} provides a visual representation of the time-lapse imaging process employed in the study.
\section{Fractal Dimension Based approaches}
Researchers can evaluate the fractal properties of data using fractal analysis, which enables them to examine the aesthetic complexities seen in different architectural forms. The inspection and comprehension of these architectural aspects is made possible by a variety of fractal analytic techniques, such as box-counting and lacunarity \cite{Lionar2021, Jimenez2012, Mandelbrot05051967}.

In earlier publications \cite{Ostwald2018}, the box-counting method was used to examine the fractal dimension (FD) of architectural structures. Notably, this study took into account architectural outline diagrams. Similar to this, \cite{Lee2021buildings} used a differential box-counting method to determine the FD of an architectural facade. The style and patterns of ancient Egyptian art have also been quantified using fractal analysis, providing information on the date and place of origin of the work \cite{Robkin2012}.

Fractal analysis has been used to investigate the complexity of architectural features of the Taj Mahal \cite{Shishin2016}. The Indian temple architecture has been the subject of other studies that have covered architectural styles, Hindu philosophy, symbolism, and FD estimate of particular structures. In these works, fractal characteristics found in temple façade and floor plans have been linked to Hindu temple design.

The 3D box-counting technique has been used to determine the fractal properties of metropolitan regions in Shenyang, China \cite{Liu2022}. Practically speaking, it can be difficult to create 3D models of real-world items or sceneries from 2D photos. As a result, it becomes more practical to analyze 2D images for fractal-like properties.

The effectiveness of convolutional neural networks (CNNs) for supervised classification tasks has led to a large increase in their use. For instance, Indonesian Temples \cite{dan2017indo} was classified using a five-layer CNN with three pooling layers. For classification purposes, transfer learning techniques using networks like AlexNet and Inception-ResNetv2 have also been used \cite{llamas2017class}.

The assessment of FD for temples with a variety of architectural styles in various parts of a country hasn't, however, received much research based on 2D photographs. One can acquire insight into the intricate designs adorning skyscraper spires by evaluating the FD values of 2D photographs.
\section{3D Reconstruction: Image Processing and Learning Based}
Standard digital cameras' inherent limitations in recording three-dimensional things through a two-dimensional medium lead to an inaccurate depiction of their physical makeup. For the sake of this discussion, points are represented by their coordinates in the X-Y plane, with $(x_1, y_1)$ standing for a position that is defined by a positive distance of $x_1$ along the X-axis and $y_1$ along the Y-axis. The Euclidean distance formula can be used to determine how far apart two points, $(x_1, y_1)$ and $(x_2, y_2)$, are from one another:
\begin{equation}
d = \sqrt{(x2 - x1)^{2} + (y2 - y1)^{2}}.
\end{equation}
Depth mapping methods \cite{basu_survay_digitalheritage2023} and numerical approximations are used to estimate the spatial profile along the X-Y plane, also known as the Z-axis, in order to produce a thorough representation of three-dimensional coordinates. In this procedure, a virtual three-dimensional system is built using the techniques covered inside a particular environmental setting.

Utilising techniques from the study of optics, the first step entails capturing the location of the distinct target object's negative or mirror reflection. According to \cite{basu_survay_digitalheritage2023}, the main component distance from the picture plane's centre is used to calculate the distance between the target item and its image. The estimation of the central axis' three-dimensional representation using a set of photos is then possible only with a positive image. Last but not least, the rotation of the $(X, Y, Z)$ coordinates is accomplished by applying three sequentially ordered rotations along the X, Y, and Z axes, which are typically denoted as $(\omega, \varphi, \kappa)$. Heritage structures frequently need more maintenance and care. We use 3D reconstruction due to the more powerful computing. As a result, we would like to talk about 3D reconstruction as a crucial component of digital reconstruction in cultural heritage. Digital technology are thought to be a dependable method for maintaining cultural heritage. Many ML and DL techniques are applied. In \cite{basu_survay_digitalheritage2023}, the paper mentioned Belhi et al. evaluated a number of DL methods for categorising and annotating heritage data as well as for filling in the gaps left by missing data.

\subsection{Image Processing based Approaches}
\subsubsection{A low cost solution for Brazillian Baroques}
In pursuit of an economical solution, many researchers explores the use of RGB Depth sensor cameras, specifically Microsoft Kinect and Intel RealSense Camera F200, to enhance imaging of sculptures like Buddha and Mozart, comparing their real-time performance and unique specifications as mentioned in \cite{basu_survay_digitalheritage2023}.
\begin{figure}[!ht]
\label{brazil}
	\centering
		\includegraphics[width=0.7\linewidth]{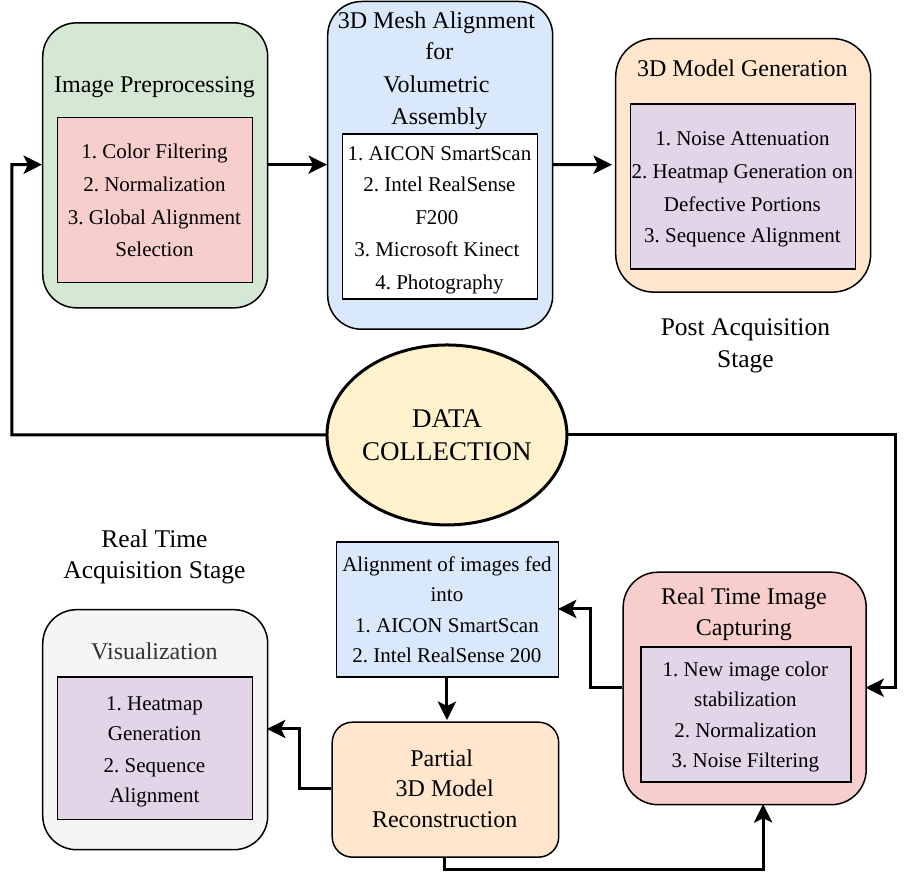}	
		\caption{Low Cost RGB-D Sensors}
\end{figure}
\begin{itemize}
  \item Stage 1: The paper utilizes a time-based 3D reconstruction approach with Microsoft Kinect and Intel RealSense cameras, capturing surface area scans at a lower resolution for real-time visualization purposes.
  
  \item Stage 2: To reduce noise, a joint bilateral or refinement filter is applied to depth maps, considering spatial and temporal information, lighting conditions, and mesh projections. Bilinear upsampling is suggested for higher resolution images.
  
  \item Stage 3: Pairwise alignment of images is achieved using the Iterative Closest Point (ICP) algorithm, which is widely used for geometrical alignment of 3D models. The proposed method covers similar images to converge different points effectively.
  
  \item Stage 4: Pulli's algorithm is employed to generate registration errors and minimize local minimum issues, offering better alignment for large datasets.
  
  \item Close-Range Photogrammetry in Yogyakarta: 1319 photos were collected from five temples, followed by keypoint extraction using the A-KAZE algorithm. A 3D reconstruction model was created using LIOP for feature matching and Sparse Point Clouds generation.
  
  \item FSSR and LST methods were employed to handle surface shape and texture using MVS as a geometrical shape generator, resulting in a higher resolution 3D reconstruction of the Yogyakarta Temples.
\end{itemize}

\subsubsection{Reconstruction on Close Range photogrammetry}
The intriguing Indonesian area of Yogyakarta has long been neglected, leaving its priceless historical sites defaced by the ravages of time and the fury of nature. These places' slow decline has tragically diminished the spiritual and cultural significance they once had, making the use of sophisticated computational techniques necessary to restore their splendour. Despite the numerous 2D images that have been taken to document these riches all around the world, they frequently fall short of capturing the minute details that make each individual artefact distinct. In light of this, \cite{basu_survay_digitalheritage2023} introduces an amazing method using up-close photography to painstakingly record the delicate details of these architectural wonders, giving their story fresh life.

\begin{itemize}
    \item The committee collected 1319 photos by visiting five temples in Yogyakarta.
    \item The effective A-KAZE algorithm, a quicker version of the Modified Local Difference Binary Descriptor, is used to extract keypoints from gathered images, enabling reliable gradient information processing.
    \item The technique introduces the use of LIOP (Local Intensity Order Pattern) for feature matching after A-KAZE keypoint extraction, followed by the building of Sparse Point Clouds to make it easier to generate a 3D reconstruction model.
    \item The study uses FSSR (Floating Scale Surface Reconstruction) and LST (Large Scale Texturing) techniques, utilising Multi-View Stereo (MVS) for precise surface shape and texture mapping, leading to a higher resolution 3D reconstruction of Yogyakarta's temple architecture \cite{basu_survay_digitalheritage2023}.
\end{itemize}

\subsection{Deep Learning based Approaches}
\subsubsection{GAN based 3D Reconstruction}
Generative Adversarial Networks (GANs) have completely changed the field of picture production \cite{goodfellow2020gans} by using adversarial game between generator and discriminator. A discriminator measures the difference between the false and real images, and GANs produce realistic images by copying features from real photos and creating counterfeit images. Deep generative models have been created for a variety of applications, including the visualisation of human faces using Deep Convolutional GANs \cite{radford2015dcgan}, thanks to this exploratory drive.

The application of GANs to heritage restoration has been quite successful. For the Borobudur temple in Indonesia, Pan et al. suggested a deep-based 3D restoration technique. With a stunning 97\% accuracy, the approach uses grayscale images to rebuild the temple's concealed areas. Deep learning (DL) and photogrammetry are used to create a 3D transparent visualisation that gives a complete picture of the temple's hidden and visible areas. There are particular difficulties in restoring parametric geometric patterns in old Islamic architecture. Hajebi created a restoration technique with a 92\% accuracy that uses entropy filters to identify damaged chunk borders and a neural network to forecast and recreate lost patterns \cite{basu_survay_digitalheritage2023}.

Over time, painted beams that have been exposed to the elements frequently deteriorate. For practically rebuilding the aged timbers of the Chinese Forbidden City, Zou et al. presented a deep-based technique. Using techniques like edge detection and semantic segmentation, the method separates the image into three components: the background, golden edges, and dragon patterns. A full image is formed by superimposing the separately restored segments.

Knyaz et al. demonstrated an image-based pipeline for dense 3D reconstruction utilising the WireNetV2 deep learning network. Unmanned aerial systems (UAS) were used to recreate Moscow's famed Shukhov Radio Tower, demonstrating the approach's superior performance in masking photos of intricate grid patterns.

The preservation and reconstruction of cultural heritage sites depend heavily on these cutting-edge methods, which include GANs, DL-based restoration, and image-based pipelines \cite{ledig2017SRGAN, wang2018esrgan}. Deep convolutional networks (CNNs) have extensively extracted descriptive features from input images. Recently, the inverse problem of image generation has been addressed in literature. Yan et al. proposed attribute-conditioned deep variational autoencoders, employing multilayer perceptrons to generate entangled representations and a course-to-fine convolutional decoder for pixel generation. However, VAE-generated images tend to be overly smooth. To overcome this, Yeh et al. \cite{basu_survay_digitalheritage2023} combined VAEs with Deep Convolutional GANs (DCGANs) to produce sharper and more realistic images.

The processing of ancient documents and artifacts presents challenges due to degradation, fading, and occlusion. Palimpsests and petroglyphs often contain partially erased or occluded portions. Traditional methods, such as segmentation and binarization, focus on visible data, leaving invisible regions unhandled. \cite{basu_survay_digitalheritage2023} mentioned Melnik et al. addressed this limitation by developing a deep network comprising a classifier and a conditional adversarial network (cGAN) capable of segmenting both visible and missing or corrupted image regions. These advanced techniques based on deep learning have significantly contributed to image generation and restoration, enabling more accurate reconstructions.

\subsubsection{CEPROQHA Project: A deep 3D reconstruction method}
The CEPROQHA Project, a project aimed at restoring Qatar's cultural heritage sites, seeks to put into practice a cutting-edge solution using sophisticated holoscopic 3D imaging for reasonably priced and high-quality outcomes. The project investigates the use of augmented reality (AR) and virtual reality (VR) to develop immersive virtual museums in the aim of improving user experiences. In addition, the suggested method in \cite{basu_survay_digitalheritage2023} capitalizes on the current trend of using deep learning (DL) and machine learning (ML), leveraging DL techniques to restore, classify, and annotate enormous amounts of missing data, while also creating standardized information mapping for cultural heritage sites. These methods improve the DL-based restoration techniques for cultural heritage preservation by utilizing generative adversarial networks (GAN) and their capacity to produce new image data, as well as an understanding of the underlying patterns within the training dataset.
\begin{figure*}[!ht]
\label{belhiworkflow}
	\centering
		\includegraphics[width=\linewidth]{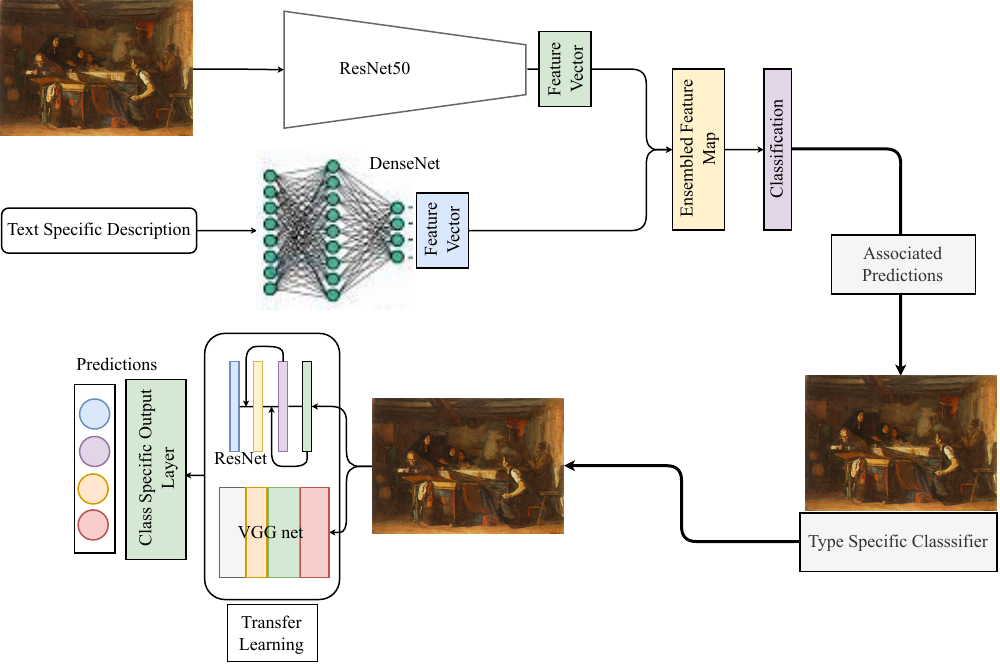}	
		\caption{Deep Reconstruction of CEPROQHA Project}
\end{figure*}
\begin{itemize}
  \item To train CNN-based classifiers, the authors use DL-based techniques and take part in data-driven cultural heritage challenges. They have amassed large datasets from the Rijksmuseum and WikiArt.
  \item With weights developed using publicly available datasets and fine-tuning utilising illustrative tuning datasets, transfer learning is used to incorporate the datasets in CNN classifiers. The topic of dataset annotation for multimodal and multitask hierarchical categorization is discussed.
  \item The study suggests multitask classification utilising a two-stage classification strategy, as well as multimodal classification using Resnet50 and dense layer feature extractors. Using deep convolutional GANs and NLP methods for ontology-based architecture reconstruction with CIDOC CRM, an end-to-end reconstruction system using GANs is also built.
\end{itemize}

\subsubsection{Deep Reconstruction of Forbidden city}
The faded coloured paintings on the historic Chinese structures have flaws like blurring, paint loss, and colour distortion. These problems are addressed by a three-step procedure described: backdrop reconstruction, restoration of golden edges, and resuscitation of detailed dragon patterns. For content segmentation and producing realistic designs, cutting-edge methods like UNET, MobileNet, and GauGAN are used. By superimposing the recreated components, this painstaking restoration procedure seeks to achieve a seamless fix. The restoration of the Region Of Interest (ROI) involves the use of computational methods and techniques from the fields of image processing and computer vision.
\begin{figure}[!ht]
\label{borobodur}
	\centering
		\includegraphics[width=0.8\linewidth]{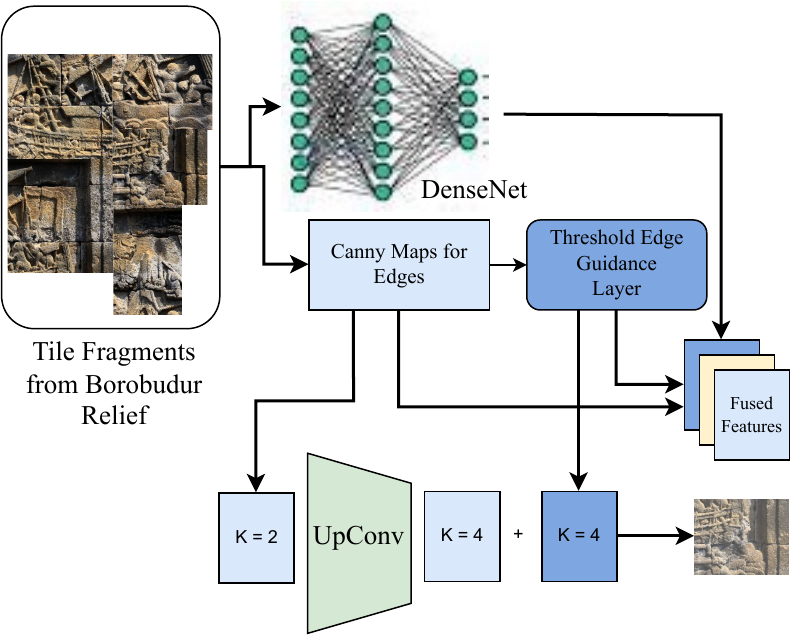}	
		\caption{GAN based reconstruction on Borobodur relief}
\end{figure}

\begin{itemize}
  \item To create a color segmentation map for the tested image, the pipeline's initial phase uses a UNET-based semantic segmentation approach that incorporates a MobileNet feature extractor. The segmentation process runs more quickly and efficiently thanks to this enhanced architecture. The next stage is mapping colors using the segmentation map that was produced.
  
  \item The corresponding step uses a perspective transformation to precisely place the chosen segmentation map on the background of black pixels.
  
  \item On the basis of the altered segmentation maps, edge detection is improved through the use of image processing techniques. Two photos are combined to produce the Complete Edge (CE) information, which is subsequently transformed morphologically by erosion and dilation to produce the Final Edge (FE) image.
  
  \item The FE pictures are extracted in the previous stage and used in the architecture for thorough background removal and edge detection. The pipeline continues by using GauGAN to superimpose the image with the dragon image while being directed by user input \cite{basu_survay_digitalheritage2023}. With the help of computational methods and GAN capabilities, this combination of photos recovers the dragon figurines' original appearance.
\end{itemize}

The overall goal of this pipeline is to restore and improve the visual quality of the old Chinese structures by combining advanced segmentation algorithms, perspective modifications, and image processing techniques.

\section{Deep Classification approaches for Cultural Heritage Sites}
Due to their diversity and complexity, cultural heritage monuments are difficult for traditional algorithms to categorise based on their intricate structures and visual content. Traditional methods fall short of capturing the complexity and depth of these architectural marvels. However, the advent of machine learning (ML) and deep learning (DL), which addressed these challenging issues head-on, has revolutionised the industry. Convolutional Neural Networks (CNNs), an example of a DL method, have shown to be successful in extracting architectural complexity and analysing pixel values in photographs. These algorithms can identify the proximity of features to the real maps by backpropagating positive feature maps and changing weights, making it easier to separate important components for additional study. This makes it possible to classify photos specifically based on their distinctive traits.
\subsection{Project Inception: CNN on heritage sites}
Llamas et al. in \cite{basu_survay_digitalheritage2023} have pioneered the use of deep learning (DL) algorithms in the study of cultural heritage data. The "INCEPTION" European project's goal is to create efficient tools for mathematical algorithms, segmentation, and Historical Building Information Modelling (HBIM) methodologies in the context of cultural heritage assets. This study extends DL to this project. The project's goal is to develop and reconstruct the 3D modelling of time-dynamic historical data so that a variety of people can access it. The project emphasises the search engine component of historical modelling rather than a strict pipeline-driven approach, aiming to deliver an immersive and well-defined visual experience by bridging social, cultural, and religious borders.

Transfer learning is used to increase the capabilities of convolutional neural networks (CNNs). This method uses a pretrained network that has been adjusted for a new dataset category on a huge dataset. The CNN learns to extract significant features and predict particular class labels by fixing the feature extractor and training only the dense layer. Transfer learning is used by Llamas et al. on a set of 10,092 photos from Flickr, with 8,000 used for training and 1,000 each for validation and testing. They use the INCEPTIONV3 model, which was previously trained on the IMAGENET dataset.

The researchers widen the concept of classification in the study in near future. They publish a collection of 10,000 photos along with related materials, using preparation techniques described in earlier publications. The designs AlexNet and ResNet are used, with ResNet's residual connections enabling better gradient flow and preventing degradation problems. The experimental settings focus on both full training and fine-tuning by freezing particular layers, taking into account different image sizes, epochs, and parameter combinations. The suggested Inception-ResNet-V2 model shows encouraging results when tested on 128 x 128 pictures for 77 epochs. The research investigates additional categories and deep learning models to evaluate their effectiveness in categorising cultural resources. The study focuses on the tension that exists in the classification task between bias-variance, learning rate, and weight decay.

In conclusion, Llamas et al. use DL-based techniques, notably CNNs, to contribute to the classification and digital documentation of cultural assets. The study analyses several architectural models, hyperparameters, and experimental setups to improve the classification of cultural heritage photographs within the INCEPTION project and shows the efficiency of transfer learning.
\subsection{Classifying disaster affected heritage sites}
Due to deterioration over time, cultural heritage structures have issues that need for improved automated classification. Let's discuss this from a new survey article \cite{basu_survay_digitalheritage2023}. Convolutional Neural Networks (CNNs) and social media data are both used in the way that Kumar et al. propose. Outliers are eliminated during data preprocessing, resulting in a dataset for classification. ResNet 50, VGG16, DenseNet 121, InceptionResNetV2, Xception, and NASNetLarge are trained CNN feature extractors. Together with ML classifiers (Logistic Regression, Support Vector Machine, Random Forest, AdaBoost, and Xception), the two models that the study focuses on are Heritage Model-1 for classifying heritage and non-heritage objects and Heritage Model-2 for classifying damage. The method has a 90\% accuracy rate for identifying early damage. The datasets used in the study come from Twitter, Google searches, and video frames and contain 6,612 heritage samples, 2,266 non-heritage samples, 836 damaged heritage samples, and 567 damaged non-heritage samples. The work makes a contribution to crisis informatics and emphasises the importance of time-dependent picture structures. Utilising DL-based methods makes it easier to identify heritage that has been harmed and allows for quick assessment. The study emphasises the necessity of automation in heritage classification to replace labor-intensive manual procedures. The multi-stage classification method advances the field and uses algorithms that are informed by data. The research's conclusions offer insightful advice for safeguarding cultural treasures.
\subsection{Deep diving into classification of ancient Mayan Glyph}
Given their historical significance, the Maya glyphs need to be preserved while still being readable and understandable. An automated method utilizing DL and ML methods was presented. Crowdsourcing and preprocessing with oversampling and geometric adjustments are used to collect the data. The number of glyph classes has increased from ten to 150.

In three different scenarios, various DL methods are tested with a focus on convolutional neural networks (CNNs):

\begin{itemize}
    \item Modifying the final layer after applying pretrained feature extractors like VGG16, ResNet18, and ResNet50 trained on ImageNet.
    \item Eliminating the final convolution block and incorporating the number of classes into a dense layer.
    \item Starting from scratch, train the entire network.
\end{itemize}

To shed light on classifier choices, the explainable AI technique GradCAM is used. GradCAM creates heatmaps that show the areas that categorization decisions should concentrate on.

It is difficult to train networks for Maya glyph categorization because there aren't many segmented glyphs available. Can et al. show that for the hieroglyph classification challenge, a sketch-specific sequential network outperforms generic sequential models (e.g., LeNet) and Residual Networks.

In conclusion, the work suggests an automated method for maintaining Maya glyphs using DL and ML methods. Following preprocessing and enhancement of crowdsourced data, CNNs are trained using various configurations. The effectiveness of a sketch-specific sequence network is highlighted in the paper, and GradCAM is used to increase the explainability of classifier decisions.
\section{Inpainting: tailing back to Traditional and Deep based methods}
An age-old method used to smoothly replace missing or damaged areas of an image while maintaining visual coherency is traditional digital inpainting. It includes choosing neighbouring picture content manually or using an algorithm as a guide to reconstruct the missing sections. Heuristic rules and custom features are used in traditional inpainting techniques to estimate and transmit the missing information. These techniques include texture generation, patch-based techniques, and diffusion-based algorithms.

By using the strength of neural networks to automatically learn the underlying patterns and structures of a picture, deep learning has, in contrast, revolutionised inpainting. Inpainting with deep learning requires training a powerful neural network on large datasets of finished images so that it can understand the nuances of the inpainting operation. Utilising the capabilities of convolutional neural networks (CNNs) or generative adversarial networks (GANs), the trained model proficiently provides realistic and coherent completions for the missing sections.

There are many benefits of deep learning-based inpainting over conventional techniques. It excels in handling complex and difficult inpainting scenarios and produces results that are both visually appealing and lifelike. The precise completion of the missing regions is guaranteed by deep learning models' capacity to acquire fine-grained image information and contextual signals. Additionally, the fact that deep learning-based inpainting is data-driven makes it easier for it to adapt to various and undiscovered inpainting jobs.
\subsection{Traditional Inpainting}
Traditional digital inpainting encompasses various approaches for replacing missing or damaged regions in images while maintaining visual coherence. These methods, such as interpolation and texture-based techniques, have been employed for small-scale inpainting tasks. Diffusion-based algorithms, such as the one proposed by Bertalmio et al., propagate information from neighboring areas to fill structured regions through an iterative process. Texture synthesis-based methods replicate sample textures from intact image regions to reconstruct the damaged areas while preserving color relations. Exemplar-based methods, introduced by Criminisi et al., select samples from the source region to fill the target region based on patch similarity. Hybrid approaches combine partial differential equation (PDE)-based and texture-based techniques for inpainting by decomposing the image into structure and texture components.

It is important to note that traditional inpainting methods have limitations in handling large missing portions and complex inpainting scenarios. Deep learning-based inpainting has emerged as a groundbreaking alternative, leveraging neural networks to learn patterns and structures from training datasets. These models, such as convolutional neural networks (CNNs) and generative adversarial networks (GANs), produce realistic and coherent completions for missing regions, surpassing the capabilities of traditional techniques.

In summary, traditional digital inpainting involves techniques like diffusion-based algorithms, texture synthesis, exemplar-based methods, and hybrid approaches. While these methods are suitable for small-scale inpainting tasks, deep learning-based inpainting using CNNs and GANs has revolutionized the field by achieving remarkable results in complex inpainting scenarios.
\subsection{Deep Inpainting}
For small-scale inpainting tasks, traditional digital inpainting techniques have been used, including interpolation, texture-based techniques, diffusion-based algorithms, texture synthesis, exemplar-based methods, and hybrid approaches. However, these methods are not suitable for handling large missing sections and complex inpainting scenarios.

The advent of neural networks, such as Convolutional Neural Networks (CNNs) and Generative Adversarial Networks (GANs), has revolutionized the field of inpainting. Deep learning-based inpainting methods leverage encoder-decoder architectures and CNNs to learn patterns and structures from training datasets, allowing them to recreate missing regions based on learned features. GAN-based techniques, such as the Wasserstein GAN and enhanced Context Encoder, generate realistic and accurate reconstructions. The PiiGAN architecture enables the generation of multiple plausible reconstructions.

In the context of 3D inpainting, methods have been developed to fill gaps and omissions in 3D point cloud data, improving the accuracy of 3D reconstructions. Additionally, forgery detection techniques have been proposed for detecting exemplar-based and diffusion-based inpainting.

Deep learning-based inpainting approaches have shown superior performance compared to traditional methods in challenging inpainting scenarios. However, it is important to note that there are numerous other works in the field of digital inpainting that are beyond the scope of this description. Interested readers can refer to more comprehensive surveys for further exploration.
\section{Object Detection Algoritms}
In computer vision applications, object detection algorithms are essential for locating and identifying objects in pictures and videos. You Only Look Once (YOLO) \cite{yolo, yolo9000, bochkovskiy2020yolov4} and Single Shot MultiBox Detector (SSD) \cite{ssdlite, liu2016ssd, huang2017ssd} are two well-known and often employed object detection methods.

A real-time object detection system called YOLO splits the incoming image into a grid and forecasts bounding boxes and class probabilities for each grid cell. It is suited for real-time applications because it processes the full image in a single pass and is renowned for its quick inference performance.

The SSD algorithms, on the other hand, are effective object detection algorithms that use a series of pre-defined anchor boxes at various scales and aspect ratios to anticipate the locations and classifications of objects. At several network layers, it anticipates object detections and performs multi-scale feature extraction.

YOLO and SSD each have their own distinctive qualities and trade-offs. While SSD strikes a balance between speed and precision, YOLO offers real-time performance but may compromise accuracy. Depending on the particular needs of the application, such as the necessity for real-time processing or high detection accuracy, one may choose among these algorithms.

Overall, YOLO and SSD offer practical and successful solutions for a range of computer vision tasks, making major advances in the field of object detection.
\subsection{YOLO: You Only Look Once}
The popular object identification technique known as YOLO (You Only Look Once) was developed by Redmon and Farhadi in 2015 \cite{redmon2015yolo}. In a single pass across the image, this one-stage method predicts object bounding boxes and class probabilities. Each grid pixel in the image is divided into a bounding box and a probability for each object type using YOLO. The YOLO concept has seen several iterations, each offering additional features and methods.

The first iteration of YOLO introduced the grid-based methodology and utilized a softmax classifier for class probabilities and a regression model for bounding box prediction \cite{redmon2016you}. YOLOv2, which was launched in 2016, enhanced the original by using a deeper network, anchor boxes for better handling of multiple objects, and a more advanced loss function.

A bigger network, an effective loss function, dilated convolutions, and the addition of the mean average precision (mAP) measure for evaluation were all included in YOLOv3 in 2018 \cite{redmon2018yolov3}. With a broader network, a new backbone network dubbed CSPDarknet53, and a robust mAP metric at various IoU thresholds, YOLOv4, which was launched in 2020, substantially improved the algorithm \cite{bochkovskiy2020yolov4}.

Additional enhancements included a bigger network, an optimized loss function, the adoption of CSPDarknet53 as the backbone network, and a training method known as "trick mode" for greater accuracy in YOLOv5, which was published in 2021 \cite{glenn_jocher_2021_4551085}. The most recent version, YOLOv8, which was launched in 2022, added a larger network, a more effective loss function, the CSPDarknet101 backbone network, and other improvements to the mAP measure and training methods \cite{glenn_jocher_2022_5568457}.

Self-driving cars, video surveillance, medical imaging, retail, and manufacturing are just a few of the industries where YOLO has found use. For further information, see Redmon (2015), Redmon (2018), and Bochkovskiy (2020) \cite{redmon2015yolo, redmon2018yolov3, bochkovskiy2020yolov4}. Its popularity and widespread use are in part due to its capability to do real-time object detection with acceptable accuracy.

Researchers are constantly working to improve YOLO's functionality, develop fresh methods, and take into account certain application needs.
\subsection{SSD: Single Shot MultiBox Detector}
Single Shot Multibox Detector (SSD), is introduced by Liu et al. to use a widely used object identification framework, in 2016. It is a single-stage object detector that uses a convolutional neural network (CNN) to predict the bounding boxes and class probabilities of objects in an image. The SSD architecture has experienced a number of developments, each bringing quite appreciable advancements.

Liu et al.'s initial implementation, SSD 300, obtained cutting-edge results by gridding the image and forecasting bounding boxes and class probabilities \cite{liu2016ssd}. Building on this achievement, Liu et al. (2017) produced SSD 512, a larger variant with a deeper CNN backbone and more default bounding boxes, to detect a wider variety of object sizes \cite{liu2016ssd512}.

In 2017, Howard et al. presented the SSD MobileNet in response to the demand for effective inference on mobile devices \cite{howard2017mobilenets}. By utilizing the MobileNet architecture, this mobile-friendly version of SSD achieves performance that is on par with SSD 512 but with noticeably quicker inference times. The SSD Refinement module was introduced by Fu et al. in 2018, and it improved the original SSD model by increasing object detection accuracy \cite{fu2018ssdrefinement}.

Model size, design, and training methods are only a few of the improvements made to the SSD framework. Improved detection over a range of item sizes is possible with larger models that have more layers and default bounding boxes, but at the cost of more computing work. The effective deployment on devices with limited resources is made possible by the incorporation of specialized architectures like MobileNet. Furthermore, exceptional performance on object detection tasks is ensured by training on benchmark datasets like PASCAL VOC and MS COCO.

The computer vision community has paid close attention to the SSD object detection framework, which offers a wide range of applications in areas including object tracking, facial recognition, and autonomous driving.
\section{Super Resolution}
\subsection{The Evolution of Super-Resolution Methods: From CNNs to GANs and Transformers}
Super-resolution techniques in the field of deep learning have progressed from relying mostly on convolutional neural networks (CNNs) to including generative adversarial networks (GANs) and, more recently, transformers. Despite the fact that CNN-based techniques were the first to use deep networks for super-resolution, they have poor generalizability and are frequently domain-specific. Domain adaptation and data augmentation have been used to try to solve this problem, but their usefulness is still limited since they lack actual generative power \cite{dong2015srcnn, dbpntpami2021, selfexsr_cvpr2015}.
\subsection{Harnessing the Power of GANs for Super-Resolution}
By including GANs in the framework, CNNs' super-resolution constraints can be solved. Using GANs with interpolation-based upscaling for super-resolution was made possible because to the groundbreaking work of SRGAN. By eliminating normalisation and using residual networks as backbones, later developments like ESRGAN and ESRGAN+ greatly enhanced the SRGAN architecture \cite{wang2018esrgan, rakotonirina2020esrgan+}. Alternative methods, such as RealESRGAN and BSRGAN, added pre-processing steps and noise injection to increase variability, but experienced difficulties keeping fine details. The use of WGAN and other upgraded GAN techniques in the context of super-resolution hasn't been fully investigated.
\subsection{The Evolution of Super-Resolution Methods: From CNNs to GANs and Transformers}
Super-resolution techniques in the field of deep learning have progressed from relying mostly on convolutional neural networks (CNNs) to including generative adversarial networks (GANs) and, more recently, transformers. Despite the fact that CNN-based techniques were the first to use deep networks for super-resolution, they have poor generalizability and are frequently domain-specific. Domain adaptation and data augmentation have been used to try to solve this problem, but their usefulness is still limited since they lack actual generative power \cite{dong2015srcnn, dbpntpami2021, selfexsr_cvpr2015}.
\subsection{Harnessing the Power of GANs for Super-Resolution}
By including GANs in the framework, CNNs' super-resolution constraints can be solved. Using GANs with interpolation-based upscaling for super-resolution was made possible because to the groundbreaking work of SRGAN. By eliminating normalisation and using residual networks as backbones, later developments like ESRGAN and ESRGAN+ greatly enhanced the SRGAN architecture \cite{wang2018esrgan, rakotonirina2020esrgan+}. Alternative methods, such as RealESRGAN and BSRGAN, added pre-processing steps and noise injection to increase variability, but experienced difficulties keeping fine details. The use of WGAN and other upgraded GAN techniques in the context of super-resolution hasn't been fully investigated.
\subsection{A Resurgence of CNNs with Attention and Transformers}
With the introduction of attention mechanisms and transformers in super-resolution, CNN-based techniques have seen a renaissance. Transformer networks in this field were made possible by SAN's introduction of channel attention \cite{sancvpr2019}. Recent works emphasise the significance of feature mixing from many levels employing attention \cite{swiniriccvw2021, zhang2022swinfir}. Examples of these works are SwinIR and SwinFIR. Layer normalisation is also used to address the problem of SR output distortions \cite{swiniriccvw2021, zhang2022swinfir}. Although uncontrolled feature mixing can produce unsatisfactory results, cross-attention has also been investigated, with NAFSSR proposing a cross-attention process. To reduce negative impacts, improved variations, like HAT-L, integrate overlapping cross-attention modules.

\chapter{Fractal Convolution based Segmentation}
\label{ref:fc}
\section{Preprocessing of the Spires}
\label{sec:preprocessing}
There are numerous methods for image processing that can be used to improve spire feature extraction. For its effectiveness in detecting edges in images, the Canny edge detector \cite{canny1986tpami} is a popular choice among these methods. To remove noise from the image, a blurring technique is first applied. The Sobel filter is then used to find the image's edges \cite{Kano1988sobel} for this. Since the Canny edge detector is a suggested non-learning-based technology for edge detection, we chose it for this study. For temple spires in particular, this filter provides better edge detection visualisation than the Sobel filter. 
\section{Fractal Dimension}
\label{sec:fractal}
The concept of \emph{dimension} refers to the minimum number of coordinates required to specify any point within Euclidean space. Consider a figure with a unit length existing in Euclidean dimension $d$. If this figure is scaled by a factor of $r$, causing the original length to reduce to $1/r$, the number of boxes needed to cover the original figure increases to $n=k.r^d$, where $k$ is a constant. By taking the logarithm on both sides of the equation, we can obtain the following expression:
\begin{equation} 
 d = \frac{ln(n)}{ln(\frac{1}{r})} + C
  \label{eq:dim}
\end{equation}
where $C=- \frac{ln(k)}{ln(\frac{1}{r})}$ is a constant for a particular $r$, such that $C \rightarrow 0$ as $d \rightarrow 0$. \\ Here, $d$ can be either an integer or a fraction. 

The fractal dimension (FD), which measures the pace at which the geometrical aspects of an item become evident at various scales, is calculated using the box-counting algorithm \cite{Suzuki2007, Jimenez2012}. The item is encircled by a grid of boxes with a side length of $s$ at each step. The FD, represented by $D$, is determined by: If $N(s)$ specifies the number of boxes the object intersects for a given length $s$, then the FD is given by:
\begin{equation}
  D = \lim_{s\to 0} \frac{ln(N(s))}{ln(1/s)} .
  \label{eq:box}
\end{equation}
FD is the slope of the line estimated by plotting $ln(N(s))$ versus $ln(1/s)$. For a 2D object, it is a value between 1 and 2.  
 A higher value in D reflects more fractal-like characteristics in the object \footnote[1]{Code available at: \url{https://github.com/rougier/numpy-100}}.
\section{Proposed Method: Fractal Convolution based Segmentation}
We use the fractal dimension (FD) as a feature extractor while using the Fractal Convolution technique. Kernels are used in convolution processes to extract features from input picture patches. In our situation, we use the fractal dimension as a mathematical metric to evaluate the richness existing in a particular patch \cite{Suzuki2007}. An architecturally rich pattern should be visually complex, with many curves and lines that represent the detailed elements found within the patch (or "tile" in the case of religious structures). We may detect the existence of curves using FD and then acquire higher values that reflect richness. 

We build an image of the exact size that is initially filled with zeros to ensure that the reconstruction keeps the same dimensions as the input image. The finished product of the full convolution process is this image. Then, each image patch is given the fractal dimension, allowing us to determine whether or not there are elaborate architectural patterns present. Higher values are produced when an image contains more curves per unit area, as in the case of dense spires, whereas lower values are produced when there are less curves, as in the case of the sky \cite{ghosh2022regenerative}.

\begin{figure}[t]
	\centering
		\includegraphics[width=0.8\textwidth]{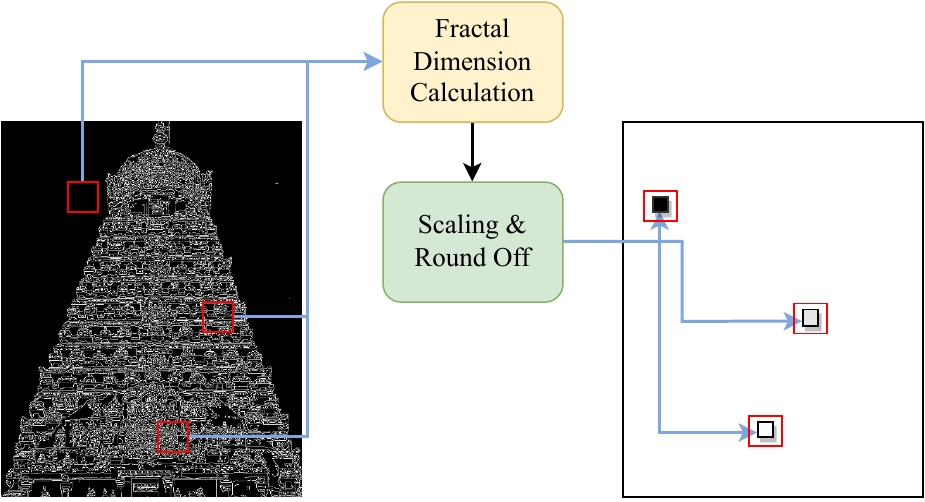}
		\caption{Graphical Representation of Fractal Convolution method}
		\label{fig:fdc}
\end{figure}

For picture reconstruction, we scale the FD value within the range of 0 to 254 by multiplying it by 127. The position within the resulting image patch is then determined by rounding off the value to make sure it is an integer. After that, the patch window is moved by one column, enabling us to repeatedly assess the richness of the succeeding parts (Figure \ref{fig:fdc}). It is important to note that the image is intentionally overlapped in order to capture the richness of each pixel, and that procedure is continued until the entire image has been processed.
The value exhibited by the FC \footnote[2]{Code available at \url{https://github.com/ecsuheritage/Frac_Dim_Digital_Heritage/blob/main/fdc.py}} method is an impression of the feature richness of the tile and present it in the middle of the reconstructed image of the same size. This value is subsequently positioned in the middle of the reconstructed image, which keeps its original dimensions. The resulting pixel will have a value near to 254, indicating a high level of richness, when the patch is filled with complicated curves. If the patch isn't as rich, though, the pixel value will be closer to 0. It should be noted that because this approach is used over the full image, reconstructing an image that is the same size as the input requires a significant amount of computer power.

Additionally, we set any negative logarithm outputs to 0 and translate the range of pixel values from 0-254 to 1-255. This method proves its superiority by assigning larger values when powerful characteristics are present in the input patch, rejecting the conventional kernel approach frequently employed in image processing filters. It exemplifies the power of the image convolution method, which combines both the mathematical and visual aspects and automatically assigns higher values to indicate the presence of rich features.
\begin{algorithm}[H]
\label{algo:fdc}
\begin{algorithmic}[1]
\State Input image
\State Rows, Cols = Input image shape
\State \textit{r = Resultant Zero image of size Rows, Cols}
\State  Input patch size
\State flag = 0
\For {\texttt{$i < rows $}}
    \For {\texttt{$j < cols $}}
        \If{$flag=0$}
            \If{$j + patch size < cols$ and $i + patch size \le rows$}
                        \State Break
            \ElsIf{$j + patch size > cols$ and $i + patch size > rows$}
                \State BGR to Gray Conversion
                \State C = Canny of the Gray image range $(50,150)$
                \State M = Mean of the Canny Image
                \State $t = round((fractaldimension(C, M)*127))$
                \If{$t \ge 0$}
                    \State $t = t+1$
                \Else
                    \State $t = 0$
                \EndIf
            \State $r[(2*i + patch size)//2, (2*j + patch size)//2] = t$
            \EndIf
        \EndIf
    \EndFor
\EndFor
\end{algorithmic}
\caption{Fractal Convolution Method}
\end{algorithm}

This rebuilt image is referred to as a rich segmentation mask. Therefore, we anticipate to see improvements in the performance of a fundamental classifier when segmented images are produced by performing bit-wise multiplication between the original image and the FC-reconstructed image.
\chapter{MosaicSlice: Data Augmentation}
\label{ref:mosiacslice}
Within the realm of our research endeavors, we have pioneered an ingenious methodology known as MosaicSlice, a captivating and avant-garde inter tile mixing strategy that transcends traditional augmentation techniques. This visionary approach, akin to a meticulously crafted artistic masterpiece, specifically caters to the unique composition of temple tiles, each harboring two distinct and mesmerizing figures.

\section{Intra-MosaicSlice: Intra tile figure mixing}
Embarking on this creative journey, we commence by meticulously selecting a solitary temple tile, a veritable tapestry that enfolds both Figure A and Figure B within its hallowed boundaries. With precision and finesse, we skillfully extract these two figures, unfurling their individual narratives onto the grand canvas of our research.

Now, we set the stage for a transformative act of fusion, an alchemical dance of visual symphony. Through a masterful interplay of horizontal flips, our artistic palette is invigorated with infinite possibilities, giving birth to an awe-inspiring ensemble of eight distinctive compositions: AB, AB', A'B, A'B', BA, BA', B'A, and B'A'. Each composition represents a harmonious synthesis of Figure A and Figure B, an intricate intermingling of their intrinsic essences having A' and B' as the horizontal flip of A and B respectively.

In our pursuit of technical excellence and visual coherence, we embark on a transformative approach to boundary manipulation, aimed at achieving a seamless fusion of two distinct figures. By leveraging advanced image processing techniques, we strive to blur the boundaries between these figures, imbuing the composition with a heightened sense of realism and aesthetic appeal.

In our endeavor to achieve boundary blurring, we employ sophisticated algorithms such as motion blur and Gaussian blur. These algorithms, rooted in mathematical principles and signal processing, introduce controlled levels of spatial smoothing along the edges where Figure A and Figure B meet. By convolving the image data with carefully designed kernel functions, we attenuate abrupt transitions and introduce a gradual blending effect.

The application of motion blur involves simulating the effect of camera or object motion during image capture, resulting in a smearing effect along the boundaries. This technique imparts a sense of dynamic movement, as if the figures are in a state of graceful transition. On the other hand, Gaussian blur employs a kernel that applies a weighted average of neighboring pixel values, resulting in a smooth, diffused transition across the boundary.

By carefully adjusting the parameters of these blurring algorithms, we strike a delicate balance between preserving the essential details of the figures and achieving a realistic fusion. The degree of blurring is meticulously calibrated to ensure that the composition retains its inherent visual coherence while effectively obscuring the distinct boundaries.

This technical approach to boundary blurring draws inspiration from the fields of computer vision and image synthesis, where researchers strive to bridge the gap between the virtual and the real. The fusion of Figure A and Figure B, facilitated by the artful application of blurring techniques, represents a convergence of computational prowess and aesthetic sensibility.

\section{Inter-MosaicSlice: Inter Tile Figure Mixing}
To unlock a new realm of possibilities and infuse our dataset with greater diversity, we embark on an innovative approach known as Inter-MosaicSlice, which involves mixing figures across multiple tiles. This approach, however, presents a unique set of challenges that must be addressed to ensure the generation of realistic and visually coherent images.
\begin{itemize}
    \item \textbf{Color Variation : } One of the challenges we encounter is the potential mismatch in color between the randomly selected tiles. These tiles may exhibit variations in lighting conditions, contrast, and even undergo different filtering processes. This can lead to a stark contrast between the figures, making it difficult to seamlessly merge them.

    To tackle this issue, we introduce an image processing-based approach that leverages the power of color filters. Specifically, we employ a sepia color filter, which not only imparts a distinct aesthetic quality but also helps to normalize the color features of the images. By applying this filter to both source images before the augmentation process, we ensure a consistent and harmonious color scheme.

    To create a realistic blended image, we adopt an averaging technique. This involves calculating the average pixel value from corresponding positions in both images and using it as the color for the new image. Additionally, we restore the brightness by applying the reverse functional derivative of the sepia filter, thereby preserving the intensity of the original colors. This results in a detailed color representation that seamlessly combines the characteristics of both source images.

    \item \textbf{Shape and Area Correlation : } Another significant challenge in inter-tile mixing lies in reconciling the differences in shape and area between the selected figures. As these figures may originate from tiles with dissimilar shapes and tile borders, it becomes crucial to establish a spatial and outer leaf correlation that maintains the ideal structure of a tile area.
    
    To address this challenge, we employ a distance-based metric using the centers of the figures from the two source images. By calculating the median and mode of the shape and correlation matrices, we ensure a balanced approach that avoids bias toward a single average value. This allows us to position the center of the resultant tile in a distance that reflects the median of the two source images.
    
    To refine the boundaries and enhance the overall visual coherence, we apply a median blur filtering technique. This helps to smooth out any irregularities in the edges, ensuring a seamless transition between the figures. Furthermore, we employ the canny edge detector to precisely define the boundaries within the designated distance. By incorporating the Euclidean distance between the center and the Harris-Corner detection line, we obtain an accurate measurement of the actual distance between the figures.
\end{itemize}
\begin{figure}[!htb]
\label{fig:mosaicslice}
	\centering
		\includegraphics[width=0.9\textwidth]{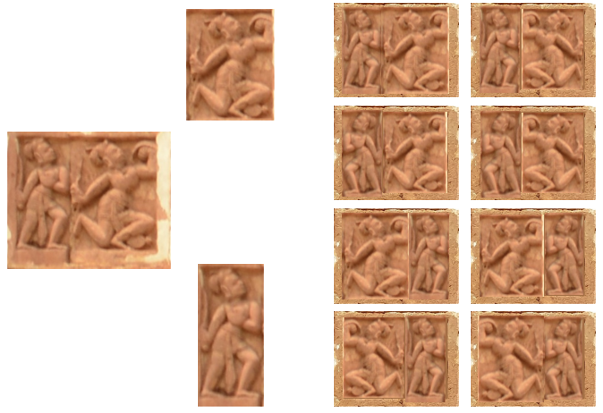}
	    \caption{A small Representation of Intra-MosaicSlice}
\end{figure}
\subsection{Harmonizing Color and Form: A Solution to Inter-Tile Mixing Challenges}
By integrating these solutions, our Inter-MosaicSlice approach overcomes the inherent challenges of color variation and shape correlation, allowing for the creation of visually appealing and realistic mosaic compositions. The meticulous application of color filters, averaging techniques, and distance-based metrics ensures a harmonious fusion of figures across multiple tiles. Through this innovative approach, we expand the possibilities of mosaic generation, pushing the boundaries of visual augmentation and artistic expression.
To address the color variation challenge, we employ the sepia color filter, which can be represented mathematically as follows:
Here, $R$, $G$, and $B$ represent the red, green, and blue color channels, respectively and $\mathcal{R'}$, $\mathcal{G'}$, and $\mathcal{B'}$ represnts normalized $R$, $G$, $B$ channel values respectively. 
\begin{equation}
\mathcal{R'} = (R \times 0.393) + (G \times 0.769) + (B \times 0.189)
\end{equation}
\begin{equation}
\mathcal{G'} = (R \times 0.349) + (G \times 0.686) + (B \times 0.168)
\end{equation}
\begin{equation}
\mathcal{B'} = (R \times 0.272) + (G \times 0.534) + (B \times 0.131)
\end{equation}

The averaging technique used to calculate the resultant color of the blended image can be expressed as:
\begin{equation}
\mathcal{R_\text{p}}(x, y) = \frac{{\mathcal{I_\text{A}}(x, y) + \mathcal{{I_\text{B}}}(x, y)}}{2}
\end{equation}
where $\mathcal{{I_\text{A}}}(x, y)$ and $\mathcal{{I_\text{B}}}(x, y)$ represent the corresponding pixels at position $(x, y)$ in the two source images.

We then focus on inversing $\mathcal{R_\text{p}}$ to get the maximised RGB intensities for the resultant image. 

\subsection{Color Normalization and Boundary Refinement: Enabling Seamless Inter-Tile Mixing}
In the quest for seamless inter-tile figure mixing, we delve into the realm of color normalization and boundary refinement. Leveraging the power of sepia color filters, averaging algorithms, and advanced image processing techniques, we bring forth a novel methodology that harmonizes color variations and establishes cohesive boundaries. By fine-tuning the color palette and blurring the edges, we create visually compelling mosaic compositions that transcend the limitations of individual tiles. This transformative approach unlocks new avenues for artistic expression and augments the potential of mosaic-based classification and analysis.
The distance metric used to determine the center of the resultant tile can be calculated using the Euclidean distance formula:

\begin{equation}
\mathcal{D} = \sqrt{{(x_2 - x_1)^2 + (y_2 - y_1)^2}}
\end{equation}

where $(x_1, y_1)$ and $(x_2, y_2)$ represent the centers of the figures from the two source images and $\mathcal{D}$ represents the distance.

The median blur filtering technique applied to refine the boundaries can be represented as:

\begin{equation}
\mathcal{R_\text{p}}(x, y) = \mathcal{M_\text{b}}(\mathcal{N_\text{n}}(x, y))
\end{equation}

where $\mathcal{N_\text{n}}(x, y)$ represents the neighboring pixels around $(x, y)$ and $\mathcal{M_\text{b}}$ calculates the median value along with the resultant pixel is represented as $\mathcal{R_\text{p}}(x, y)$.
\chapter{ProTilesGAN: Tile Genration}
\label{ref:protilesgan}
\section{Progressive Growing Architecture}
Our research makes use of a GAN-based network to make it easier to generate many tiles using the augmentation approach $D$. This network draws characteristics from the mathematical P probability distribution using a progressive upsampling method. We use a variational Progressive GAN (ProGAN), abbreviated as $G_\theta$, to build upon the architecture first presented in the ground-breaking work of Karras et al \cite{karras2017progressiveGANs}. The extension of an unbiased probability distribution is made possible by the integration of simple but extremely effective divergence measures in this specific instance of ProGAN. We derive a strategy that improves the standard and variety of generated samples by introducing a variational bound into the Generator and Discriminator architectures.
\subsection{Newly introduced Jensen-Shanon Divergence Loss}
In the context of this thesis, we want to give the generator loss function a fresh spin. We suggest adding the Jensen-Shannon (JS) divergence as an extra loss component and convexly combining it with the current Wasserstein loss. This innovative pairing encourages the creation of more varied and high-quality samples by allowing the generator to capture a wider range of distribution properties.

Let's examine their derivations to gain a complete knowledge of the mathematical underpinnings of our suggested loss functions. Consider the equations below, which highlight the essential elements:
\begin{equation}
\mathcal{L}_{\text{Wasserstein}} = \mathbb{E}_{\mathbf{x} \sim \mathcal{P}_{\text{data}}} [D(\mathbf{x})] - \mathbb{E}_{\mathbf{z} \sim \mathcal{P}_{\text{noise}}} [D(G(\mathbf{z}))]
\end{equation}
\begin{equation}
\mathcal{L}_{\text{JS}} = \frac{1}{2} \mathbb{E}_{\mathbf{x} \sim \mathcal{P}_{\text{data}}} [\log D(\mathbf{x})] + \frac{1}{2} \mathbb{E}_{\mathbf{z} \sim \mathcal{P}_{\text{noise}}} [\log(1 - D(G(\mathbf{z})))]
\end{equation}

We obtain the modified generator loss function by convexly merging the Wasserstein loss $\mathcal{L}_{\text{Wasserstein}}$ and Jensen-Shannon divergence $\mathcal{L}_{\text{JS}}$. The ability to capture both local and global distribution aspects is introduced by this enhancement, which eventually improves the quality and diversity of the created outputs.

\begin{equation}
\mathcal{L}_{\text{Combined}} = \lambda \cdot \mathcal{L}_{\text{Wasserstein}} + (1 - \lambda) \cdot \mathcal{L}_{\text{JS}}
\end{equation}

Where $\lambda$ is the convex combination coefficient, ranging between 0 and 1, $\mathcal{L}_{\text{Wasserstein}}$ represents the Wasserstein loss term, and $\mathcal{L}_{\text{JS}}$ denotes the Jensen-Shannon divergence term.
\subsection{Layer Specific Equalized Learning rate}
The equalised LR approach is used in ProGAN to make sure that the effective learning rate is balanced across layers. The ability of each layer to contribute to image synthesis may not be fully realised if a single LR value is assigned to all layers. We can assign various LR values to particular layers by taking into account how the learning rate affects the contribution of each layer, providing more precise control over the learning dynamics of the model.

Let's denote the layers in the generator network as $\{G_1, G_2, \ldots, G_L\}$, where $L$ represents the total number of layers. Similarly, for the discriminator network, we have $\{D_1, D_2, \ldots, D_L\}$.

We can provide each layer $i$ its own LR coefficient, denoted as $\alpha_i$, in order to construct layer-specific equalised LR. This parameter scales the LR for the associated layer, enabling us to adjust the learning dynamics for various network components.

The update step for each layer's parameter $i$ can be expressed mathematically as follows:
\begin{equation}
{G_i'} = {G_i} - \alpha_i \cdot \text{LR} \cdot \text{Gradient}_{G_i}
\end{equation}
\begin{equation}
{D_i'} = {D_i} - \alpha_i \cdot \text{LR} \cdot \text{Gradient}_{D_i}
\end{equation}

The gradients of the generator and discriminator with respect to their respective parameters in layer are shown here as $\text{Gradient}_{G_i}$ and $\text{Gradient}_{D_i}$. We may individually modify the effect of the learning rate on each layer thanks to the layer-specific LR coefficient $\alpha_i$. We can apply greater learning rates to layers that significantly influence picture synthesis or require more fine-tuning and lower learning rates to layers that contribute less or have already converged well by carefully choosing acceptable values for $\alpha_i$.

The network architecture, the dataset, and the desired learning dynamics will all influence the specific values chosen for the layer-specific LR coefficients. These numbers can be found by conducting extensive experimentation and analysis.

We can exert finer-grained control over the learning process by implementing layer-specific equalised LR, which could improve the production of high-quality photos by enabling various layers to more efficiently optimise their parameters.

It's crucial to remember that while this theoretical approach gives an idea for layer-specific equalised LR, its effective practical use would require empirical evaluation and fine-tuning to produce the best outcomes.
\subsection{Progressive Growing: Feature Mixing}
Feature maps are blended using linear interpolation, which is currently employed in ProGAN, by taking a weighted average between neighbouring layers. However, there are several drawbacks to linear interpolation that could result in unfavourable artefacts and jarring alterations in the images that are produced.

We can investigate the use of spline interpolation, a more sophisticated and adaptable interpolation method, to address these restrictions. Piecewise polynomial functions are used in spline interpolation to seamlessly transition between data points. Spline interpolation can offer smoother transitions and better represent the underlying structure of the feature maps by using higher-order polynomials.

Spline interpolation is a mathematical process that entails creating a collection of piecewise functions, often cubic polynomials, that pass through the input data points and meet specific smoothness requirements. These requirements guarantee that the interpolated function is continuous and has derivatives that behave properly. Smoother and more aesthetically attractive transitions are produced as a result of the interpolated function's more accurate portrayal of the underlying features.

Spline interpolation can be used to blend the feature maps between adjacent layers in the context of feature mixing in ProGAN. For each channel in the feature maps, a spline function must be built, and the weights associated with each layer's contribution must be seamlessly transitioned. It is possible to formulate the spline interpolation process as an optimisation issue,  where the objective is to find the set of polynomial coefficients that minimize the interpolation error while satisfying the smoothness constraints.

Building a collection of piecewise functions—typically cubic polynomials—that pass through a set of provided data points and meet specific smoothness requirements is known as spline interpolation. Spline interpolation can be used in the case of ProGAN to combine the feature maps between adjacent layers.

Let's use the notation $\mathbf{F}_i$ for the feature maps in the generator network, where $i$ stands for the layer index. To create a seamless transition between the two layers, the feature maps $\mathbf{F}_i$ and $\mathbf{F}_{i+1}$ will be interpolated.

We can represent each channel in the feature maps as a different curve in order to use spline interpolation. The curves corresponding to $\mathbf{F}_i$ and $\mathbf{F}_{i+1}$ should be written as $C_i(x)$ and $C_{i+1}(x)$, respectively, where $x$ denotes the point inside the feature map.
\begin{equation}
C(x) = a_3(x-x_i)^3 + a_2(x-x_i)^2 + a_1(x-x_i) + a_0    
\end{equation}

where $a_0$, $a_1$, $a_2$, and $a_3$ are the coefficients to be determined, and $x_i$ represents the position of the data point.

Additional restrictions can be placed on the derivatives of the cubic polynomials at the data points to ensure smoothness. The following set of equations results from a typical strategy of requiring continuity of the first and second derivatives:
\begin{equation}
    C_i(x_{i+1}) = C_{i+1}(x_{i+1})\\
    C'_i(x_{i+1}) = C'_{i+1}(x_{i+1})\\
    C''_i(x_{i+1}) = C''_{i+1}(x_{i+1})
\end{equation}
where $C'_i(x)$ and $C''_i(x)$ represent the first and second derivatives of $C_i(x)$, respectively.

We can find the coefficients of the cubic polynomials for both $C_i(x)$ and $C_{i+1}(x)$ by solving these equations. Using these coefficients, we may evaluate the cubic polynomials at specific locations inside the feature map to interpolate the feature maps between layers $i$ and $i+1$.

In order to provide aesthetically coherent and realistic transitions, abrupt changes and artefacts in the generated images are removed thanks to the smooth interpolation that is made possible by spline interpolation.

\chapter{Super Resolution}
\label{ref:sr}
\section{Leveraging Super Resolution with ProTilesGAN}
Due to the computational load imposed by traditional GAN architectures, producing high-resolution outputs while preserving quality is a difficulty in the field of image synthesis. We provide a novel strategy to deal with this problem that combines ProGAN and Super Resolution Generative Adversarial Network (SRGAN) architecture components.

A low-resolution (LR) image is converted into a high-quality super-resolution (SR) image by minimising a loss function along with a high-resolution (HR) counterpart in the Super Resolution Generative Adversarial Network (SRGAN) paradigm \cite{ledig2017SRGAN, wang2018esrgan, rakotonirina2020esrgan+}. We present a ground-breaking stage-wise separation technique to improve the performance of the standard SRGAN \cite{ledig2017SRGAN}.

In the past, SRGAN used two upsampling blocks in quick sequence with the generator architecture's last step, then a convolution operation. This traditional method, however, is constrained by early gradient loss since the $\mathcal{B}$ residual blocks may not amplify pixel values in the network's final stage using pixelshuffle in an ideal manner. Additionally, SRGAN \cite{ledig2017SRGAN} relies only on continuous upsampling and does not take advantage of the potential of one-stage upsampled features. By using $\mathcal{B/}2$ residual blocks followed by upsampling at each stage, our suggested stage-wise separated SRGAN design gets over these restrictions. Smaller kernel sizes capture higher-level information such item forms, relative positioning of objects, orientation, and backdrop. The choice of kernel size within these blocks is carefully studied, while larger kernel sizes capture low details such as colors, and textures.

\subsection{Enhanced Feature Blending with Convex Combination}
We use a convex combination strategy in each stage prior to upsampling to efficiently combine early convolution features with larger kernel sizes and later convolution features with smaller kernel sizes. By ensuring a harmonious blending of high- and low-level elements, this makes it possible to extract features as efficiently as possible at each stage. Let's use the notation $\mathbf{F}_{\text{prev}}$ for the feature maps from the previous stage and $\mathbf{F}_{\text{conv}}$ for the feature maps following the convolution process. The convex combination method can be expressed mathematically as follows:
\begin{equation}
    \mathbf{F}_{\text{blended}} = \alpha \cdot \mathbf{F}_{\text{prev}} + (1 - \alpha) \cdot \mathbf{F}_{\text{conv}}    
\end{equation}

where the convex combination weight, $\alpha$, which ranges from 0 to 1, is used. The contribution of the features from the previous stage and the convolution features are balanced according to this weight. We may vary the emphasis on both high-level and low-level details by modifying $\alpha$, leading to an ideal feature blending that captures the desired properties during the super-resolution phase.

\subsection{Mitigating Checkerboard Artifacts with Nearest Neighbor Upsampling}
It has been discovered that the traditional pixelshuffle operation used in the upsampling block of SRGAN causes checkerboard artefacts, or higher frequency mappings at boundary pixels, as a result of uneven convolutions that occur when lower-resolution pixels are convolved into higher-resolution space. As recommended by \cite{wang2018esrgan}, we use nearest neighbour upsampling to reduce these artefacts since it better maintains local pixel values and enables rectification through backpropagation during network training. Mathbf can be used to represent the super-resolved image as $\mathbf{SR}$ and the low-resolution image as $\mathbf{LR}$.
\begin{equation}
\mathbf{SR}(x, y) = \mathbf{LR}(\text{map}(x, y)
\end{equation}

where $(x, y)$ are the coordinates in the super-resolved image and $\text{round}(\cdot)$ denotes rounding to the nearest integer. This operation selects the nearest neighbor pixel value from the corresponding position in the low-resolution image, preserving local pixel values during upsampling. The use of nearest neighbor upsampling helps mitigate checkerboard artifacts, ensuring smoother transitions and maintaining better image quality.

\subsection{Leveraging One-Stage Upsampled Features for Improved Super-Resolution}
Each of the two phases in our stage-wise SRGAN architecture ends with an upsampling block. After the first stage upsampling, the starting block is used to ensure uniformity between the first stage and the second stage. A new feature correction technique is included in the second stage that, in addition to relying on the early features from the LR picture, also learns from the upsampled image feature space. This is noteworthy since it makes use of the features from the one-stage upsampled image. Due to the network's increased ability to capture and correct both high-level and low-level features in the upsampled image, the outcomes of super-resolution are improved. The features from the one-stage upsampled image are designated as $\mathbf{F}_{\text{upsampled}}$. The feature correction mechanism has the following mathematical form:
\begin{equation}
\mathbf{F}_{\text{corrected}} = \mathbf{F}_{\text{prev}} + \mathbf{F}_{\text{upsampled}}    
\end{equation}
where $\mathbf{F}_{\text{prev}}$ represents the early features from the LR image.
\chapter{SSTF: Self Sensitive Tile Filling}
\label{ref:sstf}
We put forth a novel strategy for temple tile restoration that integrates cutting-edge computer vision methods to automate the replacement of broken tiles. Our approach provides a comprehensive solution for producing high-quality copies of missing or damaged tiles by merging object detection, generative adversarial networks (GANs), super-resolution methods, and mathematical optimisation.
\section{Object Detection using YOLOV8}
In order to correctly identify and distinguish between broken and intact tiles inside an input image of the temple wall, we use the cutting-edge object detection algorithm YOLOV8. Given an image $I$, YOLOV8 produces a set of bounding box coordinates $B = {(x_i, y_i, w_i, h_i)}$, where $x_i$ and $y_i$ represent the coordinates of the top-left corner, and $w_i$ and $h_i$ represent the width and height of each bounding box, respectively. We train YOLOV8 to precisely detect and distinguish between damaged and unbroken tiles inside a particular image of the temple wall through meticulous annotation of the dataset in the COCO format. As a result, we are able to precisely identify the bounding box coordinates of the tile portions that are affected.
\section{Temple Tile Generation using ProTilesGAN with MosaicSlice}
We use ProTilesGAN, a GAN architecture created exclusively for temple tile synthesis, to build the missing tiles. We derive meaningful combinations of tile figures from pre-existing tiles by utilising the MosaicSlice augmentation approach. Let $T_A$ and $T_B$ stand in for two randomly chosen tiles. Eight tile combinations may be created using the MosaicSlice method: $AB$, $A'B$, $A'B'$, $A'B'A$, $BA$, $BA'$, $B'A$, and $B'A'$. The symbol for each combination is $T_C$, where $C$ stands for the figure arrangement. With this update, the dataset is enriched with a wide variety of variations that capture the detailed patterns and architectural features seen in the original tiles.
\section{Super-Resolution using StageWise Super Resolution}
We use the StageWise Super Resolution method to improve the visual appeal and amount of detail in the created tiles. The super-resolution algorithm reconstructs a high-resolution tile, $T_C^{\text{SR}}$, from a low-resolution tile, $T_C$ and a super-resolution factor, $\alpha$. This procedure has the following mathematical representation:
\begin{equation}
T_{C}^{\text{SR}} = \text{SR}(T_{C}, \alpha)    
\end{equation}
where the super-resolution function is denoted by $\text{SR}(\cdot)$. This advanced approach upscales the synthesised tiles to a higher resolution using mathematical optimisation techniques. The super-resolution technique guarantees that the replacement tiles perfectly mix with the intact tiles, faithfully replicating fine details and maintaining the overall visual coherence by taking into account prior knowledge about the underlying structure and characteristics of temple tiles.
\section{Optimal Tile Placement using Mathematical Optimization}
The created tiles must be carefully placed into the correct damaged areas as the last step. The tile placement problem is formulated as a mathematical optimisation problem. Let $T_C^{\text{SR}}$ represent the super-resolved tile corresponding to the damaged region, and $D$ represent the set of damaged tiles along with their bounding box coordinates.
\begin{figure}[!htb]
\label{fig:expidea}
	\centering
		\includegraphics[width=0.9\textwidth]{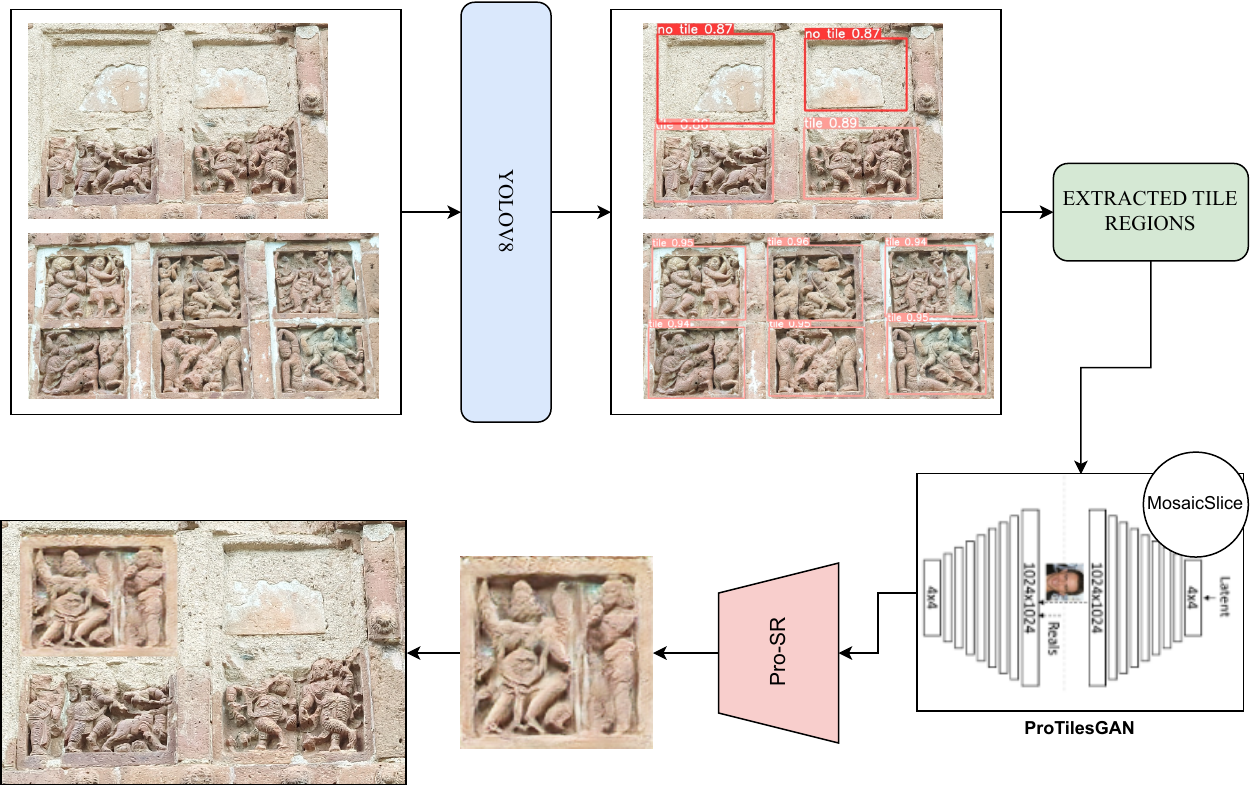}
	    \caption{Self Sensitive Tile Filling Approach}
\end{figure}
We define an objective function $f(T_C^{\text{SR}}, T_D)$ that captures the compatibility between the super-resolved tile and the damaged region. The placement problem can then be formulated as:
\begin{equation}
    T_{D}' = \arg\min_{T_D} f(T_{C}^{\text{SR}}, T_D)
\end{equation}
subject to restrictions that guarantee accurate alignment and conformance to the properties of nearby tiles. The goal of the optimisation problem is to determine where in the damaged region $T_D$ the super-resolved tile $T_C^{\text{SR}}$ should be placed.

We methodically establish the best tile arrangement by utilising mathematical optimisation approaches, taking into account numerous limitations like visual compatibility, texture consistency, and neighbouring tile attributes. By using this method, the replacement tiles will be seamlessly and tastefully integrated into the overall tile mosaic. 

\chapter{Datasets}
\label{ref:dataset}
In the history and culture of humanity, architecture has played a significant role. In this study, we set out on an enthralling journey through spires and the Bankura Terracotta Temples. We intend to reveal the complex aspects of these extraordinary structures through rephotography, deep model dataset generation, and cutting-edge algorithms. For our research, two datasets are put together. We can assess the efficacy of the Fractal Dimension (FD) approach as a metric for spire analysis using Dataset A, which consists of 100 photos. The 2,208 photos in Dataset B provide a comprehensive collection for in-depth studies. The architectural details of the Bankura Terracotta Temples are captured utilising rephotography techniques with DSLR cameras and smartphones. Dataset C from aerial photography and Dataset D from selected photography are the results of careful preprocessing. These datasets allow for the creation of high-quality tiles and super-resolution tasks when paired with cutting-edge MosaicSlice algorithms. Through this interdisciplinary project, we hope to shed light on the spires and Bankura Terracotta Temples' hidden beauty and architectural wonders, expanding knowledge of architectural analysis and preservation.

\section{Fractal Convolution Dataset:}
We have put together two different datasets, as given in Table \ref{tab:dataset}, to investigate the efficacy of fractal dimension-based convolution. Through their 2D photos, these databases seek to depict the architectural diversity of spires.
In order to fully assess the efficiency of the Fractal Dimension (FD) approach as a metric for spire analysis and classification, we used two different datasets in our study: Dataset A and Dataset B.
\subsection{Dataset A}
As a mean dataset, Dataset A has a total of 100 photos. It includes samples taken from a variety of architectural features, such as bell towers, domes, and temples. Dataset A's main goals are to determine the mean difference between photos and assess the FD method's usefulness as a valid metric for describing spire architectural aspects.

We have carefully selected 100 distinct photos from Dataset A to ensure that there is representation from a range of spire kinds. This enables us to fully assess the effectiveness and performance of the FD-based convolution technique. We may gain important insights regarding the selective power of the FD approach in separating spire changes within the dataset by quantitatively evaluating the mean difference between photos.
\begin{table}[!htb]
\caption{Mean Datasets for Fractal Convolution Method}
\label{tab:dataset}
\begin{center}
	\begin{tabular}{l|l|l}
	\toprule
	Dataset Name\hspace{0.75cm} & Temple\hspace{0.75cm} & Non Temple\hspace{0.75cm}\\
	\midrule
	\multirow{2}{*}{\textbf{Dataset A}} & \multirow{2}{*}{100} & Dome 100\\
	 & & Bell Tower 100\\
	 \midrule
	\textbf{Dataset B} & 1100 & 1108\\
\hline
\end{tabular}
\end{center}
\end{table}
\subsection{Dataset B}
With Dataset B, we have access to a larger and more complete collection of spire photos, allowing for more thorough analysis and evaluation. There are 2,208 photos in all, including both temple and non-temple spires. In order to ensure a comprehensive representation of spire types, the dataset includes a wide range of architectural elements, such as bell towers and domes.

There are several ways to use Dataset B. First of all, it is essential for teaching the classifier to do spire segmentation tasks. We can efficiently train a strong classifier capable of precisely segmenting and recognising spire regions within images by utilising the substantial data present in Dataset B. A important phase in our research workflow is this segmentation, enabling us to precisely analyze and evaluate the performance of the FD-based convolution approach.

Additionally, the initial dataset for our investigations is Dataset B. As a result, we are able to do a variety of analyses, including feature extraction, pattern identification, and comparative research, because to the extensive collection of spire photos it gives. With the use of this dataset, we can create a benchmark to compare the effectiveness of the FD method to other existing techniques or metrics frequently used in spire analysis and classification.
\section{Deep Model Dataset:}
In our quest to investigate the architectural diversity of the Bankura Terracotta Temples, we used a novel strategy that entailed the meticulous assembling of two different datasets. We intended to depict the exquisite intricacies and dynamic lighting situations of these outstanding temples by harnessing the power of rephotography and using both aerial and selected tile space approaches. Additionally, the use of state-of-the-art camera technology and our collaboration with eminent researchers assured the utmost accuracy and skill throughout the data collection process.
\subsection{Time-Dependent Rephotography}
We carried out time-dependent rephotography sessions to completely document the Bankura Terracotta Temples. Our team of photographers, concentrating on the Modonmohan temple, took pictures of numerous temple components using high-end DSLR cameras like the Canon 1500D and 3000D. These sessions covered a range of time periods, including early morning, mid-afternoon, and early evening scenarios, enabling us to photograph the temples in various lighting settings and highlight their architectural details.
\subsection{Camera Equipment and Collaboration}
Working together with experts on the Bankura Terracotta Temples, our team of photographers captured historical artefacts. We made sure the temples were completely and accurately represented by fusing our photographers' technical know-how with their historical experience. We used plenoptic vision cameras, known for their depth and stability matching capabilities, alongside DSLR cameras from Canon as well as cellphones like the Samsung M51, One Plus Nord 7, and Realme to capture a variety of viewpoints. Our team of photographers, equipped with high-quality DSLR cameras such as the Canon 1500D and 3000D.
\subsection{Preprocessing and Tile Extraction}
After gathering the data, we carefully preprocessed the data in order to extract the relevant tiles. Dataset C and Dataset D were made possible thanks to both the aerial and selected photographic sessions. While Dataset D was obtained from the selective photos, Dataset C was derived from the aerial images.
\subsection{Dataset C: Aerial Photography}
The photographs taken during the aerial photography sessions were included in Dataset C. We painstakingly picked 352 photos from the close-up aerial panels that showed off the fine features of the terracotta tiles. We used a variety of augmentation techniques, such as sharpening, horizontal flipping, brightness adjustment between 1-3\%, and histogram normalisation, to increase the dataset's diversity and quality. A rich and varied portrayal of the aerial tiles was made possible by these additions, which caused Dataset C to be expanded to 2470 photos.
\subsection{Dataset D: Selective Photography}
Images obtained by selective photography were included in Dataset D. Due to the selective technique, these photographs showed finer details, enabling us to precisely capture the intricate characteristics of the terracotta tiles. We carefully cropped 1400 photos from Dataset D to remove them from the tile boundaries. We used augmentation methods similar to those used for Dataset C, such as horizontal flipping, brightness correction, sharpening, and histogram normalisation, to produce a sizable collection of 7390 photos for Dataset D.
\section{MosaicSlice Algorithms}
Our sophisticated MosaicSlice algorithms, Intra MosaicSlice (\(\mathcal{M_{IN}}\)) and Inter MosaicSlice (\(\mathcal{M_{IT}}\)), both used Datasets C and D as inputs. These methods were developed to produce the X8 and X24 datasets, which stand for aesthetic and statistical significance, respectively. These datasets would be used as training material for our ProTilesGAN and StageWise Super Resolution models, allowing us to produce high-quality tiles, carry out super-resolution tasks, and carry out an extensive architectural examination of the Bankura Terracotta Temples.
\begin{table}[!ht]
\caption{Description of Dataset C, D, and E}
\label{tab:my-table}
\begin{tabular}{@{}c|c|c|c@{}}
\toprule
Dataset &
  Augmentations &
  \begin{tabular}[c]{@{}c@{}}Augmentation \\ Type\end{tabular} &
  \begin{tabular}[c]{@{}c@{}}Number of \\ Samples\end{tabular} \\ \midrule
\multirow{4}{*}{C} &
  No &
  NA &
  352 \\
 &
  Yes &
  \begin{tabular}[c]{@{}c@{}}Brightness (1-3\%), \\ Sharpening, \\ Histogram Equalization, \\ Horizontal Flip, Gaussian Noise, \\ Blurring\end{tabular} &
  2,470 \\
 &
  Yes &
  Intra MosaicSlice \(\mathcal{M_{IN}}\) &
  19,760 \\
 &
  Yes &
  Inter MosaicSlice \(\mathcal{M_{IT}}\) &
  59,280 \\
  \midrule
\multirow{4}{*}{D} &
  No &
  NA &
  1,400 \\
 &
  Yes &
  \begin{tabular}[c]{@{}c@{}}Brightness (1-3\%), \\ Sharpening, \\ Histogram Equalization, \\ Horizontal Flip, Gaussian Noise, \\ Blurring\end{tabular} &
  7,390 \\
 &
  Yes &
  Intra MosaicSlice \(\mathcal{M_{IN}}\) &
  59,120 \\
 &
  Yes &
  Inter MosaicSlice \(\mathcal{M_{IT}}\) &
  1,77,360 \\
  \midrule
  E & Yes & \begin{tabular}[c]{@{}c@{}}Brightness (1-3\%),\\ Sharpening,\\ Histogram Equalization,\\ Horizontal Flip, Gaussian Noise,\\ Blurring\end{tabular} & 1325 \\
 &  &  & \\
 &  &  &  \\ \bottomrule
\end{tabular}
\end{table}
\section{Dataset E: Region Detector}
We provide a second dataset made up of 1325 wall photos in order to identify the spaces between the current tiles and the places where tiles have been damaged or changed. Both regions with intact tile and tile-free portions are shown in these photos. Using the labelImg tool, we painstakingly annotate the regions of interest in the dataset to aid in the training process. The annotated dataset is then divided into 147 images for the validation set and 1178 photos for the training set.

During the dataset development process, we use a variety of augmentation strategies, similar to those used in Dataset C and D, to improve the resilience and generalisation abilities of our object identification model. These augmentation techniques aid in diversifying the dataset and provide our model with a comprehensive understanding of the diverse visual patterns and variations encountered in real-world scenarios.

In conclusion, our efforts to collect datasets of the Bankura Terracotta Temples involved the use of rephotography methods, professional cooperation, and cutting-edge camera equipment. Complex preprocessing and augmentation techniques, along with aerial and selective photography, ensured the capture of minute details and changing lighting conditions. The generated datasets will be useful tools for increasing tile production, super-resolution, and architectural study in the field of Bankura Terracotta Temples, further improved by our ground-breaking MosaicSlice algorithms.

\chapter{Qualitative and Quantitative Evaluation}
\label{ref:exp}
\section{Fractal Convolution: Subjective and Statistical}
We use the strength of two different dataset types in our effort to make the Fractal Dimension (FD) a useful statistic. Samples from the Temple, Dome, and Bell Tower architectures are included in Dataset A, often known as the Mean Dataset. We meticulously test this dataset and calculate the mean difference between 100 photos \footnote[3]{The dataset is available at \url{https://github.com/ecsuheritage/Frac\_Dim\_Digital\_Heritage}}. The varied spheres of dome and bell tower architectures are covered by Dataset B, a fusion of Temple and Non-Temple spires. This dataset serves as both the starting point for our investigations and the training set for segmentation classifiers.

We seek to visually validate the effectiveness of the Fractal Dimension (FD) approach by analysing the experimental findings and tabulating them. This section explores datasets A and B in-depth, illuminating our experimental methods and findings.
\vspace{-.5cm}
\subsection{Preprocessing:}
\label{sec:exppre}
We use a number of preprocessing methods to both datasets A and B in our pursuit of fractal dimension-based analysis. First, we handle the pixel-level noise that exists in the photos and requires the use of a noise removal technique. We also convert the images from three-channel representations to single-channel grayscale in order to support Fractal Convolution. 

We give the tables \ref{tab:example} to objectively assess the effects of preprocessing on dataset A. These tables support our assertion that Fractal Dimension is a useful parameter for spire detection by offering thorough insights into the efficiency of our preprocessing methods. We specifically give the FD(O) and FD(P) values, which stand for the original and preprocessed images' respective Fractal Dimensions.
\begin{table}[!htb]
    \centering
     \caption{Mean Fractal Dimension of Original (O) and Preprocessed (P)}
    \begin{tabular}{c|c|c|c|c|c}
    \toprule
    Monument Name & \multicolumn{2}{c}{\textbf{FD}} & \multicolumn{2}{c|}{\textbf{FD}} & Temple Name \\
    & FD(O) & FD(P) & FD(O) & FD(P) & \\
    \cmidrule{1-6}
    Taj Mahal & 1.6438 & \textbf{1.6875} & 1.6327 & \textbf{1.8426} & Kandariya\\
    Dome of the Rock & 1.5147 & \textbf{1.7107} & 1.6066 & \textbf{1.7962} & Murudeshwara\\
     St. Peter's Basilica & 1.6259 & \textbf{1.6598} & 1.5258 & \textbf{1.7735} &  Kedareshwara\\
     La Giralda & 1.5954 & \textbf{1.6362} & 1.5132 & \textbf{1.7455} & Dakshineshwar\\
     Borobudur Stupa & 1.6425 & \textbf{1.8947} & 1.7754 & \textbf{1.7683} & Shyam Rai\\
     \bottomrule
    \end{tabular}
    \label{tab:example}
\end{table}
\subsection{Fractal Convolution Method:}
\begin{figure}[!htb]
\label{fig:expfc}
	\centering
		\includegraphics[width=.6\textwidth, height =.5\textheight]{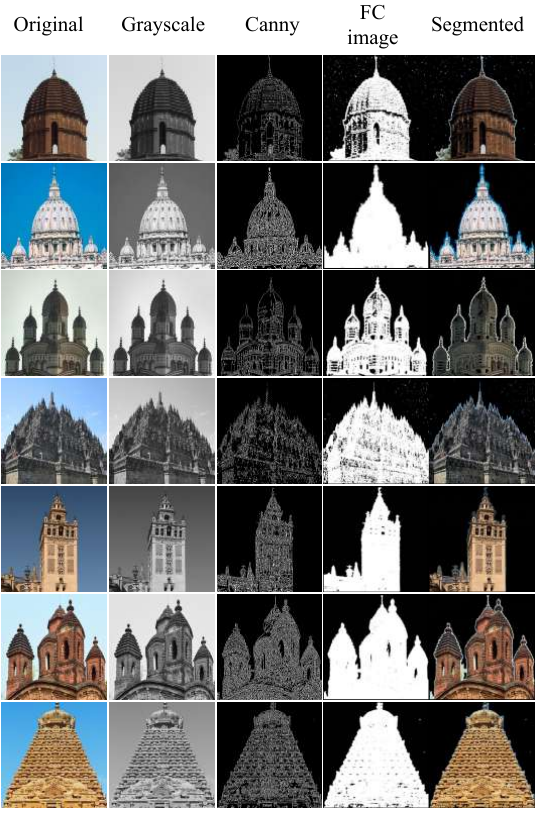}
	    \caption{Fractal Convolution Method}
\end{figure}

We use Fractal Convolution-based procedures to put our suggested strategy into practise. We display the findings as visuals by scaling the data within a given range. Some of the experimental results of the approach covered in Section this article are shown in Figure \ref{fig:expfc}. The usefulness of our approach in creating precise segmentation masks for input images is then demonstrated using image segmentation techniques.
\vspace{-0.5cm}
\subsection{Classification:}
We use a Convolutional Neural Network (CNN) classifier to demonstrate the efficacy of our segmentation method. Our tests used a standard CNN model as the classifier. The results from dataset B, which is a superset of dataset A, are shown in Table \ref{tab:classification}. When compared to earlier advances, the classification performance exhibits a substantial improvement of about \textbf{8\%}.
\begin{table}[!htb]
\caption{CNN classification on Original and Fractal Convolution Dataset}
\label{tab:classification}
\begin{center}
	\begin{tabular}{ccccc}
	\hline 
 Data set & Category & Recall & Precision & F1 Score \\
 \hline
 B & ~~Original~~ & ~~83.4543~~ & ~~84.7519~~ & ~~84.0981~~ \\
 B & ~~Segmented~~ & ~~\textbf{91.4285}~~ & ~~\textbf{92.9782}~~ & ~~\textbf{92.1968}~~\\
\hline
\end{tabular}
\end{center}
\end{table}

\noindent \emph{Experimental Setting:} We carefully set the lower bound and upper bound of the Canny edge detection technique to 50 and 150, respectively, during the preprocessing phase. To get the best edge detection results, these values were picked out empirically. 

We selected a $8 \times 8$ patch size for the fractal convolution algorithm. After extensive testing, it was discovered that this patch size offered the ideal compromise between keeping little features and guaranteeing accurate reconstruction. Notably, higher patch sizes, such $20 \times 20$ and $33 \times 33$, produced notable alterations, possibly affecting the reconstruction's quality.

We used a completely connected network and the Adam optimizer for network training. Crossentropy, a popular loss function for multiclass classification, was chosen as the loss function. We trained the network for a total of 100 epochs while closely monitoring the test accuracy to ensure robust model convergence. The model with the highest test accuracy was kept for later review and analysis.
\section{ProTilesGAN: A visual and tabular comparison with Benchmarks}
\begin{figure}[!htb]
\label{fig:exppro}
	\centering
		\includegraphics[width=0.9\textwidth]{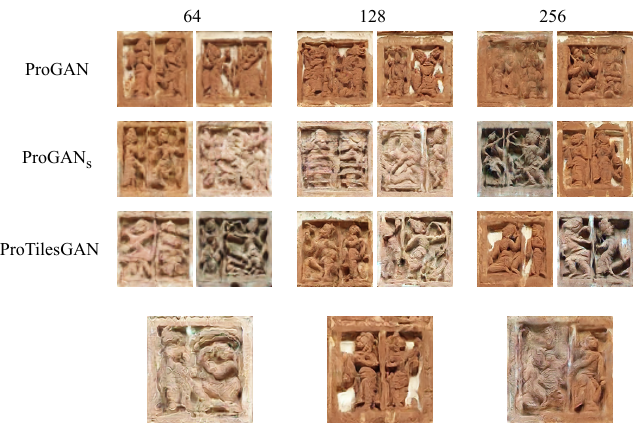}
	    \caption{Tile Generation Comparison with baseline appraoches}
\end{figure}

Since 2021, the introduction of progressive growing in Generative Adversarial Networks (GANs), led by the ground-breaking ProGAN, has been a significant advancement in the field. We have used the power of this progressive growth methodology in our study to create a customised version that is geared towards the production of delicate tiles. We added unique methods to the progressive growth paradigm, such as automatic learning rate equalisation from prior layer activations, sophisticated interpolation, and normalised smooth blending. Additionally, by penalising negative gradients through a gradient penalty and adding a feature-based L1 loss into the GAN's critique, our divergence techniques have given the gradients a newfound coherence. The ProTilesGAN method has successfully filled the gap in the area of metrocity and revolutionised tile generating. We demonstrate a qualitative comparison of ProGAN, $\text{ProGAN}_s$, and ProTilesGAN's abilities to generate tiles in Figure \ref{fig:exppro}. The three stages of the progressive growing generation are represented by graphics that range in size from 64 to 256. The model gradually updates the generation by smoothly mixing the data using an interpolation technique with a scale factor of 2 as it learns the fundamental characteristics for each size.
Despite the fact that ProGAN and $\text{ProGAN}_s$ have acceptable generation quality, our observations show that there is still potential for improvement, notably in tile generation. When trained on standard data, ProGAN finds it difficult to capture the nuanced differences found in the tiles and frequently creates tiles with just one sort of complexity. Additionally, it regularly fails to produce cohesive combinations of features in the internal and boundary regions of the tile. Even if the use of MosaicSlice increases the variety, it still has issues creating tiles of sizes 128 and 256. The ProGAN algorithm seems to be particularly good at creating 2D non-geometric characteristics, such as flat paper drawings, but falls short when it comes to capturing the depth map included in the Bankura temple tiles.

In comparison, the results from our ProTilesGAN progressive generation technique are encouraging. The gradients are made more susceptible to the depth map through the probability distribution of generation by adding smooth blending, enhanced interpolation methods, and a KL loss. The critic penalises the depth map produced by the generator using a feature-based mean squared error (MSE). As a result, the generator grows skilled at creating a more accurate depth map to fool the reviewer in the intensely competitive game. Notably, as we explore further convolutions, ProTilesGAN demonstrates great success in producing tiles with clearly defined human forms, displaying abrupt alterations along boundaries, internal areas, and smooth blending. Table \ref{tab:protiles} present a tabular comparison of FID and SSIM between ProGAN.

\begin{table}[]
\centering
\caption{A tabular Comparison of FID amd SSIM with bechmark GAN architectures}
\label{tab:protiles}
\begin{tabular}{@{}c|ccc|ccc|c@{}}
\toprule
\multirow{2}{*}{Method} & \multicolumn{3}{c|}{FID} & \multicolumn{3}{c|}{SSIM} & \multirow{2}{*}{MosaicSlice} \\ \cmidrule(lr){2-7}
 & 64 & 128 & 256 & 64 & 128 & 256 &  \\ \midrule
DCGAN & 21.64 & 22.85 & 20.64 & 0.43 & 0.39 & 0.47 & No \\
ProGAN & 20.75 & 20.23 & 18.75 & 0.42 & 0.41 & 0.58 & No \\
ProGAN\_s & 17.28 & 12.69 & 10.47 & 0.47 & 0.56 & 0.65 & Yes \\
ProTilesGAN & \textbf{14.65} & \textbf{10.42} & \textbf{7.64} & \textbf{0.59} & \textbf{0.62} & \textbf{0.77} & Yes \\ \bottomrule
\end{tabular}
\end{table}

Our research has demonstrated the utility of divergence and progressive growth algorithms in the context of tile production. ProTilesGAN has advanced tile production by capturing fine details and creating examples that are visually attractive thanks to its new methodologies and thorough understanding of depth maps. We anticipate future research initiatives that push the limits of generative models and encourage a deeper understanding of the aesthetic complexities contained in cultural heritage tiles as we continue to pursue the advancement of this discipline.


\section{Progressive Stage SR: A subjective Comparison}
The generative ability of ProTilesGAN is hampered by hardware restrictions in our system. We use a stage-wise progressive super resolution method to get around this restriction. In contrast to the traditional super resolution frameworks, our suggested architecture not only solves super resolution but also includes an interpolation-based dissimilar component. This innovative framework prioritises the integration of information through a weighted combination and places a heavy emphasis on the generative capabilities of the network, allowing for a more targeted treatment of high and low-level features.
\begin{figure}[!htb]
\label{fig:expck}
	\centering
		\includegraphics[width=0.9\textwidth]{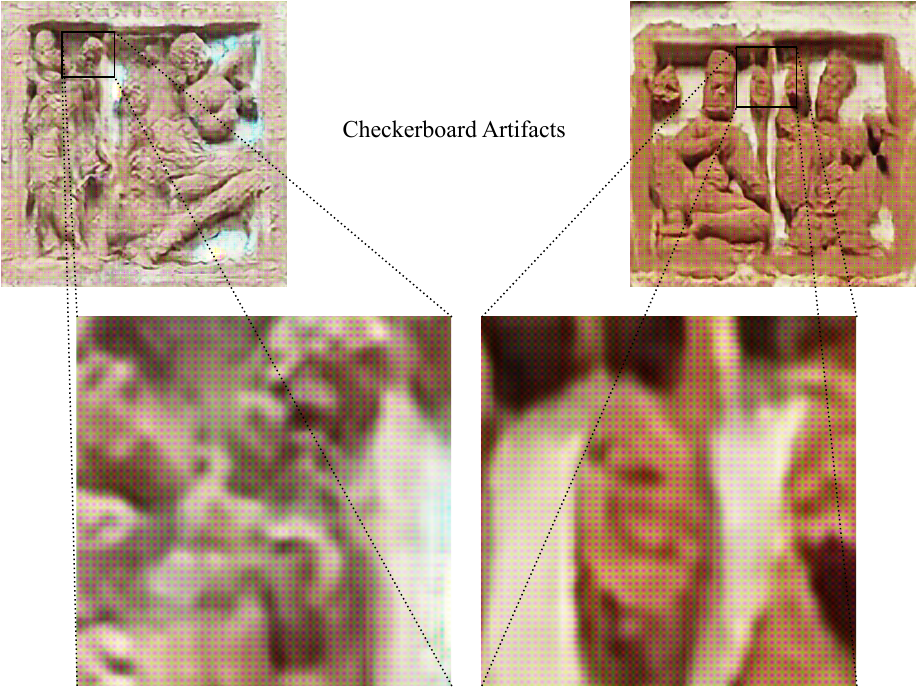}
	    \caption{Checkerboard Artifacts}
\end{figure}
Checkerboard artefacts, a frequent practical result in super resolution when convolution maps overlap, are visible in Figure \ref{fig:expck}. When the network uses deconvolution or transposed convolution to scale up the initial gradient maps, this effect is particularly obvious. We analyse this problem in-depth within the framework of ESRGAN in our study and provide a solution by including an interpolation stage followed by deconvolution. The appearance of checkerboard artefacts within our network is effectively normalised by this adaptation. As a result, our stage-SR technique shows that artifact-free generation is possible, with progressive growth taking place in two stages, each of which includes two scale factors. 

It is important to note that checkerboard artefacts develop as a result of the upscaled image's unequal information distribution, creating a grid-like pattern with alternating high- and low-frequency components. Super-resolved images' perceived realism and visual quality can both be severely impacted by these artefacts. We want to reduce these artefacts and improve the overall quality of the generated images by utilising interpolation and deconvolution techniques.

After dealing with the issue of checkerboard artefacts, our research is moving on to enlarging the image size using Progressive Stage-SR. Our SR technique seeks to increase the image size from $256$ to a stunning $1024$ using a two-stage progressive generation procedure, whereas ProTilesGAN's generating capacity is restricted to $256$ due to hardware limitations. In order for the approach to learn the complex distribution patterns of tiles inside the dataset, we train it using Dataset D.

We use bicubic downsampling with a scale factor of $4$ to lower the resolution of each image in Dataset D during the training phase. The stage-SR module then creates a Super Resolution (SR) image from the Low-Resolution (LR) image that has been downscaled. By computing the loss and comparing it to the matching High-Resolution (HR) ground truth image, we can ensure the accuracy of the SR image and direct the training process in the direction of the intended goal. Our technique attempts to push the limits of image size and improve the overall resolution of created images by utilising the power of progressive generation and training on a diverse dataset. This results in a sophisticated and scalable solution for image expansion and enhancement.
\begin{figure}[!htb]
\label{fig:expsr}
	\centering
		\includegraphics[width=0.9\textwidth]{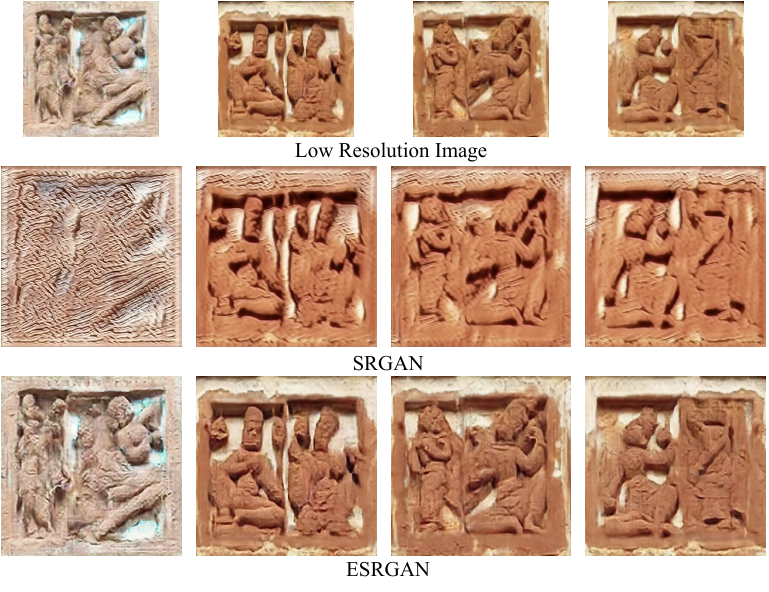}
            \includegraphics[width=0.9\textwidth]{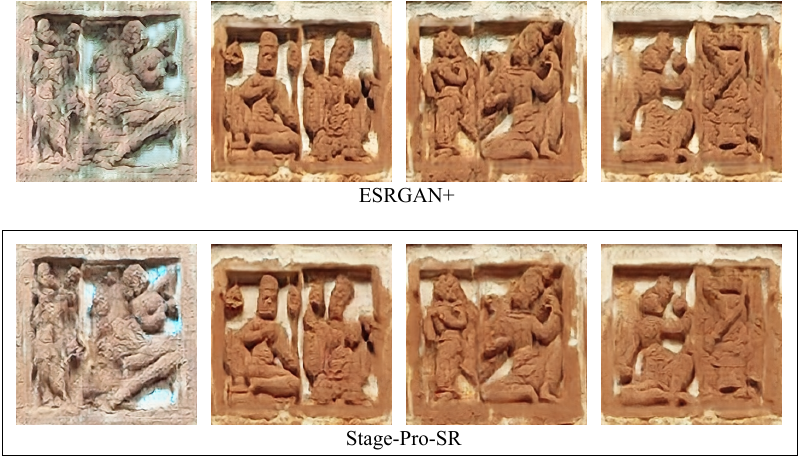}
	    \caption{SRGAN generation Comparison}
\end{figure}
We undertake a comparative study with three baseline approaches, SRGAN \cite{srgan_cvpr2017}, ESRGAN \cite{wang2018esrgan}, as shown in Figure \ref{fig:expsr}, in order to thoroughly assess the effectiveness of our methodology. To guarantee equitable evaluation circumstances before the comparison, all algorithms are trained using Dataset D.

When examined, we find that SRGAN partially transforms the brownish samples, but it severely fails to produce the varied distribution patterns found in the tiles. On the other hand, subjective brightness and contrast performance are improved by both ESRGAN and ESRGAN+. However, these techniques still have trouble achieving the needed generating capability while successfully removing the lingering problem of checkerboard artefacts. They perform better than SRGAN, but are still outperformed by our Stage-Pro-SR approach, which successfully eliminates the negative impacts of checkerboard artefacts while also avoiding bias in terms of brightness and contrast. Notably, our method outperforms ESRGAN and ESRGAN+ in terms of producing tiles with finer features (as seen in the fourth sample in Figure \ref{fig:expsr}), all the while ensuring minimal distortion in the appearance of tiles, a limitation apparent in ESRGAN+. 

Our approach stands out as a formidable solution as a result of this thorough comparative research, showing greater performance in producing tiles with increased richness, complexity, and authenticity while minimising common artefacts and maintaining the underlying properties of the information.
\section{Region Detector: YOLOv8}
\begin{figure}[!htb]
\label{fig:expyolo_vis1}
	\centering
		\includegraphics[width=0.9\textwidth]{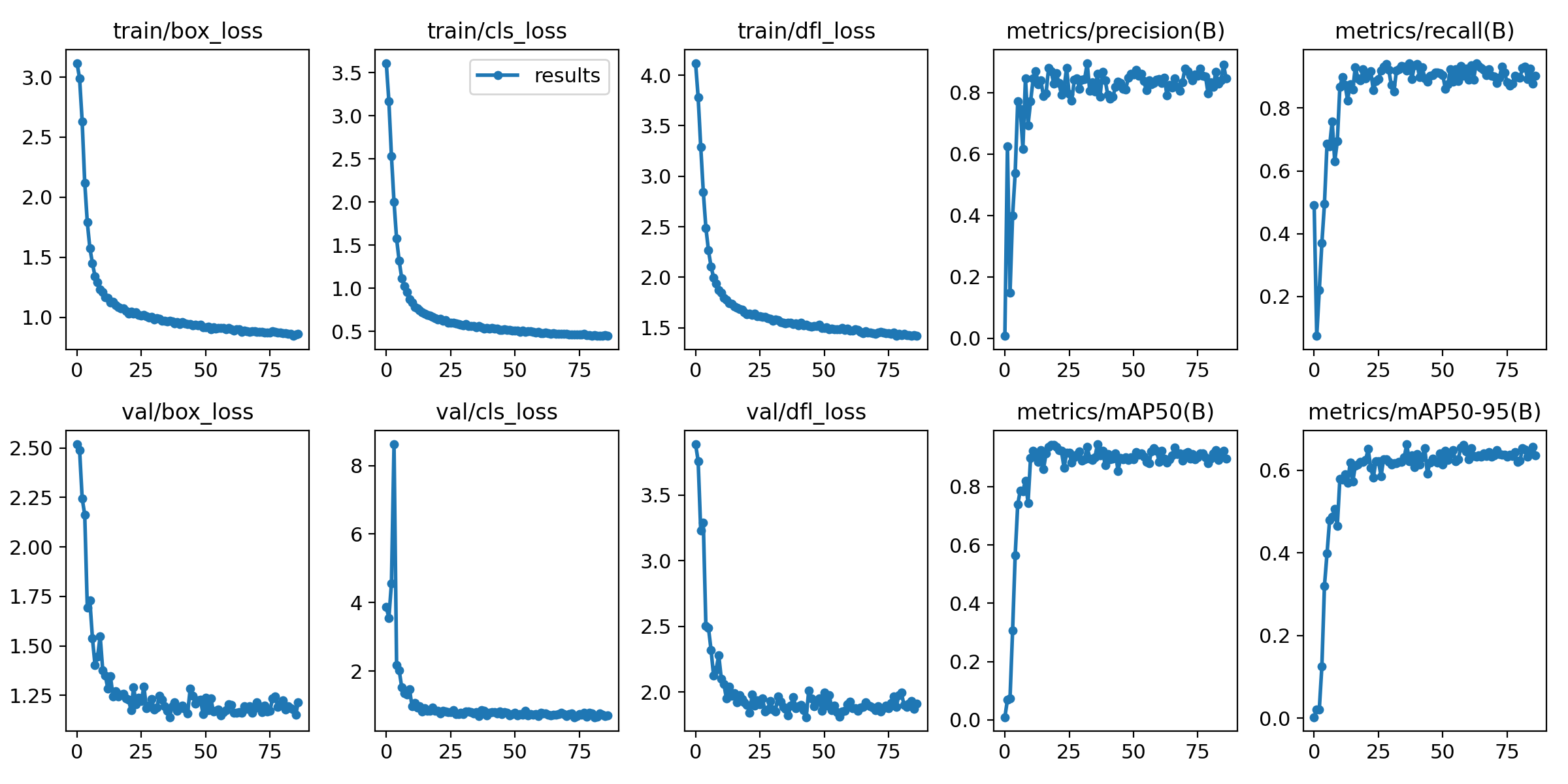}
	    \caption{Calculation of Loss and Mean Average Precision for Dataset E}
\end{figure}

We use the cutting-edge object detection technology YOLOv8 \cite{redmon2015yolo} from Ultralytics in order to broaden the scope of our research and explore the world of automation. This technique is useful for analysing Dataset E, 
\begin{table}[!htb]
\centering
\caption{A Tabular presentation on classification report on test set of Dataset E}
\label{tab:exp_cm}
\begin{tabular}{@{}c|ccc|@{}}
 & No Tile & Tile & Background \\ \midrule
No Tile & \textbf{0.91} & 0.00 & 0.31 \\
Tile & 0.01 & \textbf{0.99} & 0.69 \\
Background & 0.08 & 0.01 & 0.00 \\ \bottomrule
\end{tabular}
\end{table}
which consists of annotated walls showing the locations of portions with and without tiles. The algorithm skillfully recognises and understands the distinctive regions based on the annotations provided by the annotator by utilising the notion of region of interest.
\begin{figure}[!htb]
\label{fig:expyolo}
	\centering
		\includegraphics[width=0.9\textwidth]{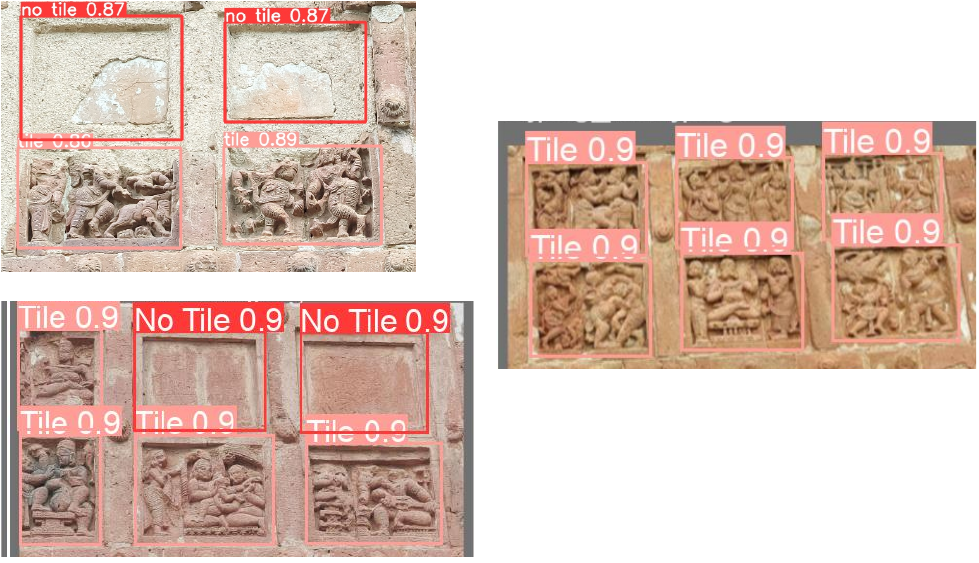}
	    \caption{Tile and No Tile Region Detetion}
\end{figure}
The confidence idea is explained in Figure \ref{fig:expyolo} and \ref{tab:exp_cm}, which also displays the places that a rigorously trained YOLOv8 model identified. Figure \ref{fig:expyolo_vis1} presents an overview of minimization of training, validation loss alongwith the improvement in Mean Average Precision for Dataset E, region detector dataset. We use the YOLOv8 model's nano variant to get the best performance by custom-tuning it for our dataset. We chose a batch size of 16 and an image size of 384 x 640 while keeping efficiency and effectiveness in mind, which allowed us to achieve a healthy balance between computing demands and precise detection skills.
With the help of this YOLOv8 integration into our research framework, our system is now capable of quickly and precisely identifying and differentiating between regions with and without tiles. By utilising YOLOv8, we improve the accuracy and efficiency of our pipeline and establish the groundwork for cutting-edge automated methods in the field of heritage restoration.

\section{SSTF: Qualitative Results}
We successfully complete our work under the Seamless Super Tile Filling (SSTF) paradigm by combining the ground-breaking results obtained through the merger of ProTilesGAN, Stage-Pro-SR, and YOLOv8 region identification. We arrange a smooth series of actions that perfectly capture the core of our study methodology using a complex flow diagram.

We begin the procedure by generating a large number of diverse samples from ProTilesGAN until a preset threshold, capturing the essence of temple wall aesthetics, is attained, drawing on the essential principle of picture normalisation. The created tiles are then subjected to super resolution using our ground-breaking Stage-Pro-SR network, bringing their visual quality to previously unheard-of heights. We carefully choose the best model based on the wall tile normalisation to provide the best outcomes, giving us the luxury of a variety of models that suit various preferences. Additionally, we present a global model that is capable of producing flawlessly synchronised tiles across a wide range of normalisations. We advocate a selective model approach that allows for rapid image generation using a hierarchical paradigm, giving users freedom.

This painstakingly constructed framework exemplifies our unwavering pursuit of excellence in the field of heritage restoration. It is supported by cutting-edge research and strengthened by a combination of advanced algorithms. The coordinated interaction of ProTilesGAN, Stage-Pro-SR, and YOLOv8 region detection within the SSTF framework establishes a new standard for automated tile generation, taking the discipline to previously unattainable levels of accuracy and effectiveness.
\begin{figure}[!htb]
\label{fig:expsstf}
	\centering
		\includegraphics[width=0.9\textwidth]{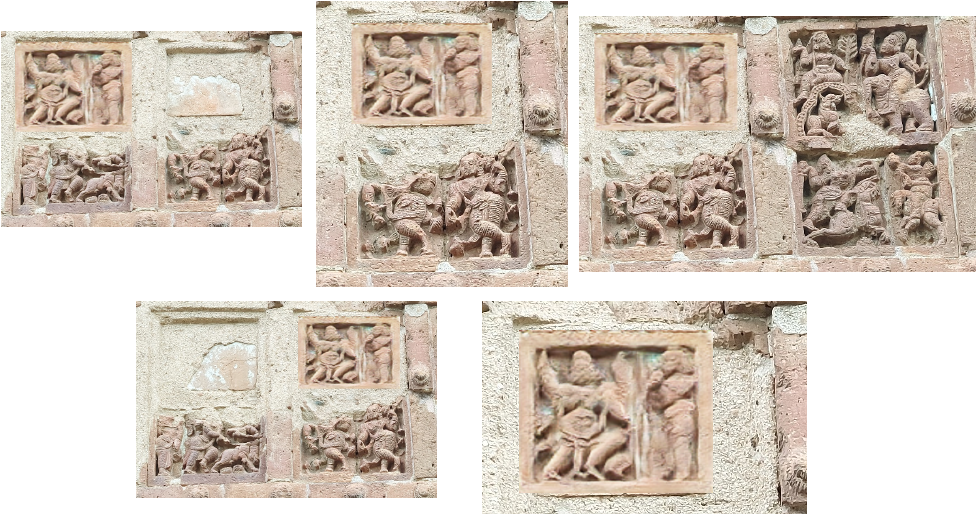}
	    \caption{Self Sensitive Tile Filling Approach}
\end{figure}
In order to distinguish between discrete regions of tiles and non-tiles within the collected image, our methodology seamlessly incorporates a sophisticated region recognition algorithm after the tile generation process. With the use of this useful information, we go on to properly resize the chosen image and align its distribution with the tile wall's. This painstaking scaling procedure creates the framework for producing incredibly genuine tiles inside the non-tile zones.

We use cutting-edge image processing methods to skillfully blend the corners of the enlarged image with the surrounding non-tile region in order to produce a seamless integration. A critical stage in creating outcomes that are aesthetically pleasant and visually coherent is this rigorous blending procedure. As shown in Figure \ref{fig:expsstf}, our method seamlessly integrates the synthesised tile with the preexisting architectural context by replacing the created super-resolved tile into the empty tile space.

Our methodology demonstrates the height of innovation in the field of tile restoration by utilising cutting-edge algorithms and cutting-edge research techniques. Our persistent dedication to attaining unmatched realism and integrity in the restoration of legacy tiles is best demonstrated by the seamless interaction of region identification, accurate scaling, and intelligent blending.
\chapter{Conclusion and Future Work}
\label{ref:conclusion}
\section{Future Works}
There is yet untapped potential for additional research despite the great breakthroughs made in the field of digital restoration through the use of computerised tools and techniques. Here, we present a collection of insightful future study directions that are quite valuable:
\begin{enumerate}
    \item The critical task of 3D reconstruction is at the centre of the first phase of heritage restoration. We can improve the effectiveness and precision of this approach by incorporating automated point cloud extraction techniques, such those suggested by Grilli et al. (2019). We adopt the idea of Graph Convolution as an inspiration in place of human statistical methods. Even though it requires a lot of processing, this method presents a potential way to divide up point cloud data. Applying approaches to optimise the extraction process allows for scalable implementation on a greater scale. Additionally, the combination of 3D localization signals gleaned from point clouds with 2D object detection techniques like YOLO (You Only Look Once) enables effective segmentation and subsequent reconstruction of items inside the image.
    \item We intend to improve the reconstruction process in the second step by including Internet of Things (IoT) detectors. These cutting-edge sensors, connected via fast internet, record the geometry of the surrounding structure. As previously shown, we may use deep learning techniques for software-free reconstruction by leveraging these architectural elements. This methodology is demonstrated to be a useful and efficient replacement for conventional software-based strategies in digital heritage reconstruction. Utilising the positional data gathered from the 3D sensors, classification and point prognosis methodologies can be used to generate identity-based mapping after 3D reconstruction.
    \item Future research will focus on researching other modalities and enhancing discrimination of digital heritage data in order to improve temple spire segmentation and classification. In order to speed up processing, we will maybe turn the fractal convolution procedure into a tensor operation. During the regeneration process, we will also put a lot of effort into conserving the complexity of the architecture and capturing minute details. The objective is to improve digital restoration techniques and aid in the thorough preservation of cultural assets.
    \item Utilising GAN-based digital inpainting techniques to fill internal gaps based on contextual data obtained from the surroundings is another potential approach. To provide fresh samples, StoryGAN uses time sequence storytelling. We may use the created succession of tiles to seamlessly fill damaged regions through digital inpainting by training the model on a set of tiles that convey a story. We may use deep learning scenarios to expand this strategy into the world of 3D models, as seen in recent works like SimpleRecon.
    \item Data of great significance is derived from the training sets thanks to the exceptional capacity of computer vision algorithms. The integration of StoryGAN architecture with diffusion models which have transformed the creation of high-definition images, for HD photos and 3D diffusion—was investigated in the previous direction. We can recreate the environment of restoration by smoothly inpainting damaged areas depending on the storytelling cues offered by the tiles by analysing the marriage of StoryGAN with diffusion-based DreamFusion architecture. These approaches could also make it easier to recreate historic buildings in 3D with high-quality. These methods can also be applied to real-time inpainting employing video inpainting methodology, providing a cutting-edge way to repair cracked tiles in a structure.
    \item While SSTF offers a comprehensive solution, digitally filling tiles necessitates additional image processing capabilities and user participation. Therefore, the creation of an interactive beta testing platform is crucial in order for us to assess the effectiveness of GANs and obtain insightful feedback. Diffusion models have also shown the potential to outperform conventional GANs by producing better images, therefore exploring this field has promise. We can improve our methods and guarantee the preservation of cultural treasures with unmatched accuracy and visual appeal by embracing these developments.
\end{enumerate}
Through these future research directions, we aim to unlock the true potential of digital restoration, ushering in a new era of preservation and revitalization for our cultural heritage.
\section{Conclusion}
In order to preserve cultural property, the article emphasises the value of machine learning-based restoration techniques while minimising manual intervention. With a focus on dataset collecting, 3D reconstruction, digital inpainting, and classification algorithms, large-scale datasets and a variety of automated procedures are investigated. The efficacy of these methods is assessed using genetic algorithms and statistical fitness functions. We provide a cutting-edge method termed fractal convolution that makes use of the fractal dimension to improve image segmentation and classification of religious structures. Furthermore, our all-encompassing SSTF strategy provides a comprehensive answer for locating and repairing damaged tile areas in temples. We effortlessly incorporate cutting-edge stage-SR network to reconstruct and resuscitate these architectural treasures with the finest precision and artistry, fueled by the astonishing powers of ProTilesGAN and enhanced by the brilliant MosiacSlice technology.

This study effectively eliminates the need for manual intervention by highlighting the value of machine learning-based restoration techniques in the field of cultural heritage preservation. We have explored dataset collecting, 3D reconstruction, digital inpainting, and classification methodologies using a wide range of large-scale datasets and automated procedures. The effectiveness of these methods has been assessed using genetic algorithms and statistical fitness functions.

Our ground-breaking use of fractal convolution, which harnesses the power of fractal dimension to provide unmatched image segmentation and classification of religious structures, is one example. Furthermore, the thoroughness of our SSTF approach demonstrates our unwavering dedication to finding and repairing damaged tile areas in temples. We effortlessly combine state-of-the-art stage-SR networks to breathe fresh life into these architectural wonders with unmatched accuracy and artistry, propelled by the extraordinary capabilities of ProTilesGAN and enhanced by the brilliant MosiacSlice method.

As we bid farewell to this thesis, may its conclusions serve as an inspiration and a spark to future academics and enthusiasts who are passionate about heritage preservation. May the fabric of our combined efforts continue to unfurl, creating a knowledge and innovation legacy that resonates through the ages.

\bibliographystyle{IEEEtran}

\end{document}